\documentclass[10pt,twocolumn,letterpaper]{article}

\usepackage{cvpr}              

\definecolor{cvprblue}{rgb}{0.21,0.49,0.74}
\usepackage[pagebackref,breaklinks,colorlinks,allcolors=cvprblue]{hyperref}

\usepackage{graphicx}
\usepackage{amsmath}
\usepackage{amssymb}
\usepackage{booktabs,caption}
\usepackage{multirow}
\usepackage{bm}
\usepackage{dsfont}
\usepackage[linesnumbered,ruled,vlined]{algorithm2e}
\usepackage{makecell}
\usepackage[flushleft]{threeparttable}
\usepackage{enumitem}
\usepackage[accsupp]{axessibility} 
\usepackage{algorithmic}
\usepackage[dvipsnames]{xcolor}
\definecolor{MyBlack}{HTML}{323A45}

\graphicspath{{figures/}}
\usepackage[pagebackref,breaklinks,colorlinks]{hyperref}

\def\etc{\emph{etc.\ }}

\usepackage[capitalize]{cleveref}
\crefname{section}{Sec.}{Secs.}
\Crefname{section}{Section}{Sections}
\Crefname{table}{Table}{Tables}
\crefname{table}{Tab.}{Tabs.}

\def\paperID{99} 
\def\confName{CVPR}
\def\confYear{2025}

\title{NTIRE 2025 Challenge on Event-Based Image Deblurring: Methods and Results}

\author{Lei Sun$^*$ \and
Andrea Alfarano$^*$ \and
Peiqi Duan$^*$ \and
Shaolin Su$^*$ \and
Kaiwei Wang$^*$ \and
Boxin Shi$^*$ \and
Radu Timofte$^*$ \and
Danda Pani Paudel$^*$ \and
Luc Van Gool$^*$ \and
Qinglin Liu \and
Wei Yu \and
Xiaoqian Lv \and
Lu Yang \and
Shuigen Wang \and
Shengping Zhang \and
Xiangyang Ji \and
Long Bao \and
Yuqiang Yang \and
Jinao Song \and
Ziyi Wang \and
Shuang Wen \and
Heng Sun \and
Kean Liu \and
Mingchen Zhong \and
Senyan Xu \and
Zhijing Sun \and
Jiaying Zhu \and
Chengjie Ge \and
Xingbo Wang \and
Yidi Liu \and
Xin Lu \and
Xueyang Fu \and
Zheng-Jun Zha \and
Dawei Fan \and
Dafeng Zhang \and
Yong Yang \and
Siru Zhang \and
Qinghua Yang \and
Hao Kang \and
Huiyuan Fu \and
Heng Zhang \and
Hongyuan Yu \and
Zhijuan Huang \and
Shuoyan Wei \and
Feng Li \and
Runmin Cong \and
Weiqi Luo \and
Mingyun Lin \and
Chenxu Jiang \and
Hongyi Liu \and
Lei Yu \and
Weilun Li \and
Jiajun Zhai \and
Tingting Lin \and
Shuang Ma \and
Sai Zhou \and
Zhanwen Liu \and
Yang Wang \and
Eiffel Chong \and
Nuwan Bandara \and
Thivya Kandappu \and
Archan Misra \and
Yihang Chen \and
Zhan Li \and
Weijun Yuan  \and
Wenzhuo Wang \and
Boyang Yao \and
Zhanglu Chen \and
Yijing Sun \and
Tianjiao Wan \and
Zijian Gao \and
Qisheng Xu \and
Kele Xu \and
Yukun Zhang \and
Yu He \and
Xiaoyan Xie \and
Tao Fu \and
Yashu Gautamkumar Patel \and
Vihar Ramesh Jain \and
Divesh Basina \and
Rishik Ashili \and
Manish Kumar Manjhi \and
Sourav Kumar \and
Prinon Benny \and
Himanshu Ghunawat \and
B Sri Sairam Gautam \and
Anett Varghese \and
Abhishek Yadav \and
}

\begin{document}

\maketitle

\let\thefootnote\relax\footnotetext{$^*$ L. Sun (lei.sun@insait.ai, INSAIT, Sofia University ``St. Kliment Ohridski''), A. Alfarano, P. Duan, S. Su, K. Wang, B. Shi, R. Timofte, D. P. Paudel, and L. Van Gool were the challenge organizers, while the other authors participated in the challenge.\\
Appendix~\ref{sec:teams} contains the authors' teams and affiliations. \\
NTIRE 2025 webpage: \url{https://cvlai.net/ntire/2025/}. \\
Code: \url{https://github.com/AHupuJR/NTIRE2025_EventDeblur_challenge}.}

\begin{abstract}
This paper presents an overview of NTIRE 2025 the First Challenge on Event-Based Image Deblurring, detailing the proposed methodologies and corresponding results. The primary goal of the challenge is to design an event-based method that achieves high-quality image deblurring, with performance quantitatively assessed using Peak Signal-to-Noise Ratio (PSNR). Notably, there are no restrictions on computational complexity or model size. The task focuses on leveraging both events and images as inputs for single-image deblurring. A total of 199 participants registered, among whom 15 teams successfully submitted valid results, offering valuable insights into the current state of event-based image deblurring. We anticipate that this challenge will drive further advancements in event-based vision research.
\end{abstract}

\section{Introduction}
\label{sec:introduction}

Traditional camera output frames with relatively long exposure time in a fixed framerate.
In contrast, event cameras, a kind of neuromorphic sensor, asynchronously capture pixel-wise intensity changes with high temporal resolution \cite{Gallego22pami}, and have been applied in various fields such as computational imaging~\cite{sun2024unified,event_based_frame_interpolation_with_ad_hoc_deblurring,Event_Based_Fusion_for_Motion_Deblurring_with_Cross_modal_Attention,messikommer2022multi,sun2025low}, human pose estimation~\cite{chen2022efficient}, depth estimation~\cite{lu2024sge,muglikar2021esl}, image segmentation~\cite{zhang2023multi,alonso2019ev}, \etc.

In recent years, significant efforts have been dedicated to event-based image restoration. Among various tasks, event-based image deblurring has gained the most attention, as the high temporal resolution of event cameras provides valuable priors for motion deblurring~\cite{sun2024unified,event_based_frame_interpolation_with_ad_hoc_deblurring,Event_Based_Fusion_for_Motion_Deblurring_with_Cross_modal_Attention}. Notably, these methods operate under the assumption that input images and events are spatially aligned—a condition that applies to all approaches discussed in this paper.

In conjunction with the NTIRE 2025 Workshop on New Trends in Image Restoration and Enhancement, the Event-Based Image Deblurring Challenge was organized. The objective is to develop a network architecture or solution that effectively integrates events and images to enhance image deblurring performance. We hope that this challenge will serve as a starting point for promoting event-based image enhancement on a broader stage and contribute to the thriving development of the event-based vision community.

This challenge is one of the NTIRE 2025~\footnote{\url{https://www.cvlai.net/ntire/2025/}} Workshop associated challenges on: ambient lighting normalization~\cite{ntire2025ambient}, reflection removal in the wild~\cite{ntire2025reflection}, shadow removal~\cite{ntire2025shadow}, event-based image deblurring~\cite{ntire2025event}, image denoising~\cite{ntire2025denoising}, XGC quality assessment~\cite{ntire2025xgc}, UGC video enhancement~\cite{ntire2025ugc}, night photography rendering~\cite{ntire2025night}, image super-resolution (x4)~\cite{ntire2025srx4}, real-world face restoration~\cite{ntire2025face}, efficient super-resolution~\cite{ntire2025esr}, HR depth estimation~\cite{ntire2025hrdepth}, efficient burst HDR and restoration~\cite{ntire2025ebhdr}, cross-domain few-shot object detection~\cite{ntire2025cross}, short-form UGC video quality assessment and enhancement~\cite{ntire2025shortugc,ntire2025shortugc_data}, text to image generation model quality assessment~\cite{ntire2025text}, day and night raindrop removal for dual-focused images~\cite{ntire2025day}, video quality assessment for video conferencing~\cite{ntire2025vqe}, low light image enhancement~\cite{ntire2025lowlight}, light field super-resolution~\cite{ntire2025lightfield}, restore any image model (RAIM) in the wild~\cite{ntire2025raim}, raw restoration and super-resolution~\cite{ntire2025raw} and raw reconstruction from RGB on smartphones~\cite{ntire2025rawrgb}.

\section{NTIRE 2025 Event-Based Image Deblurring Challenge}

The goals of this challenge include: (1) promoting research in the area of event-based image deblurring, (2) facilitating comparisons between various methods, and (3) providing a platform for academic and industrial participants to engage, discuss, and potentially establish collaborations. This section delves into the specifics of the challenge, including the dataset, challenge phases and evaluation criteria.

\subsection{Dataset}
The HighREV dataset \cite{event_based_frame_interpolation_with_ad_hoc_deblurring} is used for both training and evaluation in this challenge. It consists of 1,771 sets of blurry images, corresponding events, and sharp images for training. Additionally, 421 sets are provided as validation data during the development phase, ensuring a comprehensive benchmark for assessing model performance.


\subsection{Tracks and Competition}
The aim is to obtain a network design capable to produce high-quality results with the best performance measured by PSNR for event-based image deblurring.

\medskip
\noindent{\textbf{Challenge phases }}
Participants were given access to training images from the HighREV dataset. During the validation phase, they could use 421 images from the validation set for model tuning. In the test phase, evaluation was performed on 271 images from the test set. To ensure a fair assessment, the ground-truth images for the test phase remained hidden from participants throughout the challenge.

\medskip
\noindent{\textbf{Evaluation protocol }}
Since the aim of this challenge is to foster the development of accurate event-based image deblurring networks, PSNR and SSIM on the 271 testing images are used as the quantitative evaluation metrics. A code example for calculating these metrics is available at \url{https://github.com/AHupuJR/NTIRE2025_EventDeblur_challenge}. The code of the submitted solutions and the pre-trained weights are also available in this repository.

\section{Challenge Results}

Table \ref{tab:rank} shows the final rankings and test results of the participated teams. The implementation details of each team can be found in Sec.\ref{sec:methods_and_teams}, while team member information can be found in Appendix \ref{sec:teams}. IVISLAB achieved the first place in terms of PSNR, followed by MiVideoDeblur and 404NotFound as the second and third place, respectively.

\begin{table}[t]
    \centering
    \begin{tabular}{l|c|cc}
    \hline
    Team & Rank & PSNR (primary) & SSIM  \\ \hline
    IVISLAB & 1 & 42.79 & 0.9196 \\  
    MiVideoDeblur & 2 & 42.70 & 0.9281 \\
    404NotFound & 3 & 42.09 &  0.9300 \\  
    Give\_it\_a\_try & 4 & 40.37 &  0.9234 \\  
    BUPTMM & 5 & 40.21 & 0.9179 \\  
    WEI & 6 & 39.46 & 0.9171 \\  
    DVS-WHU & 7 & 39.26 &  0.9101 \\  
    PixelRevive & 8 & 39.12 &  0.9112 \\  
    CHD & 9 & 38.56 & 0.9055 \\  
    SMU & 10 & 38.30 & 0.9047 \\  
    JNU620 & 11 & 37.63 & 0.9019 \\  
    colab & 12 & 36.84 &  0.8962 \\  
    CMSL & 13 & 31.81 &  0.8900 \\  
    KUnet & 14 & 29.42 & 0.8600  \\  
    Group10 & 15 & 25.93 &  0.8200 \\  
    \hline
    \end{tabular}
    \caption{Results of NTIRE 2025 Event-Based Image Deblurring Challenge. PSNR and SSIM scores are measured on the 271 test images from HighREV dataset. Team rankings are based primarily on PSNR.}
    \label{tab:rank}
\end{table}

\subsection{Participants}
\label{sec:participants}
The challenge attracted 199 registered participants, with 15 teams successfully submitting valid results.

\subsection{Main Ideas and Architectures}
\label{sec:main_ideas}
Throughout the challenge, participants explored various innovative techniques to improve deblurring performance. Below, we summarize some of the key strategies employed by the top-performing teams.

\begin{enumerate}
    \item \textbf{Hybrid architectures demonstrated strong performance,} with all top-3 teams utilizing a combination of transformers and convolutional networks. This approach leverages global features extracted by transformers alongside local features captured by convolutional layers, both of which contribute to effective event-based image deblurring. Besides, both spatial and channel attention mechanisms play a crucial role in enhancing overall performance.
    \item \textbf{Pretrained weights matters.} The winning team, IVISLAB, leveraged a backbone model initialized with pretrained weights from ImageNet, demonstrating the advantages of transfer learning in event-based image deblurring.
    \item \textbf{Cross-modal fusion proves beneficial.} Several teams adopted EFNet~\cite{Event_Based_Fusion_for_Motion_Deblurring_with_Cross_modal_Attention} and REFID~\cite{event_based_frame_interpolation_with_ad_hoc_deblurring,sun2024unified} as a baseline model to fuse features from the event and image branches.
    \item \textbf{Effective training strategies.} Both the second and third-place teams employed progressive learning techniques during training. Additionally, the winning team utilized a large patch size ($512\times512$), which contributed to improved performance.
    \item \textbf{Incorporating a novel Mamba-based architecture}. Integrating features from both image and event modalities is crucial for enhancing the reconstruction quality of event-based deblurring methods. Team DVS-WHU introduced an innovative Mamba-based architecture to achieve more effective fusion.

\end{enumerate}

\subsection{Fairness}
To maintain fairness in the event-based image deblurring challenge, specific rules were implemented, primarily regarding the datasets used for training. Participants were permitted to use external datasets for training. However, incorporating the HighREV validation set, whether sharp or blurry images, was strictly prohibited, as this set served to evaluate the overall performance and generalizability of the models. Additionally, the use of HighREV test blurry images for training was not allowed. On the other hand, employing advanced data augmentation techniques during training was considered an acceptable practice.


\section{Challenge Methods and Teams}
\label{sec:methods_and_teams}

\subsection{IVISLAB}
\begin{figure}[t]
    \centering
    \includegraphics[width=1.\linewidth]{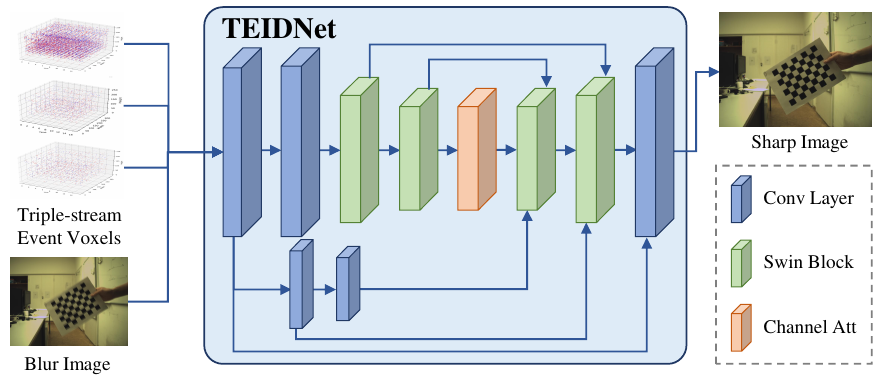}
    \caption{The model architecture of TEIDNet, proposed by Team IVISLAB.}
    \label{fig:dernet}
\end{figure}

To achieve image deblurring, team IVISLAB introduces the Triple Event-stream Image Deblurring Network (TEIDNet). As depicted in Figure~\ref{fig:dernet}, TEIDNet converts consecutive events into event voxels at three temporal scales to perceive motion information from blur images and capture fine edges for reconstructing clear images. Furthermore, TEIDNet integrates Shift Window Attention and Channel-Wise Attention blocks to capture local and global contexts, thereby enhancing deblurring accuracy.

\subsubsection{Network Architecture}
TEIDNet adopts an encoder-decoder architecture to process images and triple-stream event voxels, aiming to estimate the deblurred image. Specifically, when deblurring the image at frame $t$, TEIDNet considers that the long-term event stream surrounding frame $t$ can aid in motion perception. Therefore, it voxelizes the event data from frame $t-T_l$ to frame $t+T_l$ into a $b$-bin event voxel $V_{l,t}$. 
Simultaneously, since the short-term event stream around frame $t$ can help reconstruct high-frequency textures, TEIDNet voxelizes the event data from frame $t-T_s$ to frame $t+T_s$ into a $b$-bin event voxel $V_{s,t}$. Furthermore, to mitigate color artifacts by leveraging higher-resolution motion information near the current frame, TEIDNet voxelizes the event data from frame $t-T_m$ to frame $t+T_m$ into a $b$-bin event voxel $V_{m,t}$.
Subsequently, the event voxels $V_{l,t}$, $V_{s,t}$, and $V_{m,t}$, along with the blur image $I_b$, are concatenated and fed into the network. To effectively fuse the features from the image and event voxels, TEIDNet employs convolutional layers to generate fused feature representations.
The network then utilizes a dual-branch encoder. The first, a complex branch extracts high-level semantic information from the fused features by leveraging shift window attention to capture local context and channel-wise attention blocks to capture global context. The second, a simple branch utilizes convolutional layers to capture fine-grained details from the fused features.
Next, TEIDNet’s decoder integrates multiple shift window attention blocks to fuse and upsample the features extracted by the dual-branch encoder. Finally, convolutional layers are employed to predict the deblurred image $I_t$.

\subsubsection{Loss Function}
To train TEIDNet, they define a reconstruction loss $\mathcal{L}_r$ for the estimated deblurred image $I_{t}$ as follows:
\begin{equation}
\mathcal{L}_r = \lambda_1 \, \text{L}_1(I_{t}, I_{t}^{gt}) + \lambda_2 \, \text{L}_2(I_{t}, I_{t}^{gt})
\end{equation}
Here, $\lambda_1$ and $\lambda_2$ are coefficients that balance the loss terms. The function $\text{L}_1(\cdot,\cdot)$ represents the mean absolute error, while $\text{L}_2(\cdot,\cdot)$ denotes the mean squared error. The term $I_{t}^{gt}$ refers to the ground truth image at frame $t$.

\subsubsection{Implementation Details}
TEIDNet is implemented using PyTorch on four Nvidia L20 GPUs. During training, a batch size of 16 is utilized, with input data dimensions of $512 \times 512$ pixels. The network weights are optimized over 1000 epochs using the AdamW optimizer, with an initial learning rate set to $2 \times 10^{-5}$. A cosine annealing scheduler is employed to decay the learning rate progressively. In addition, they take the checkpoint with good performance and perform a second finetune. To mitigate overfitting, data augmentation techniques such as random flipping and rotation are applied. They also initialize the backbone network parameters using weights  pre-trained on ImageNet. The specific coefficients and parameters are defined as follows: number of bins $b = 7$, long-term temporal window $T_l = 5$, medium-term temporal window $T_m = 1$, short-term temporal window $T_s = 0$, and loss function weights $\lambda_1 = 1$, $\lambda_2 = 1$.

\subsection{MiVideoDeblur}
\begin{figure}[tbp]
    \centering
    \includegraphics[width=0.49\textwidth]{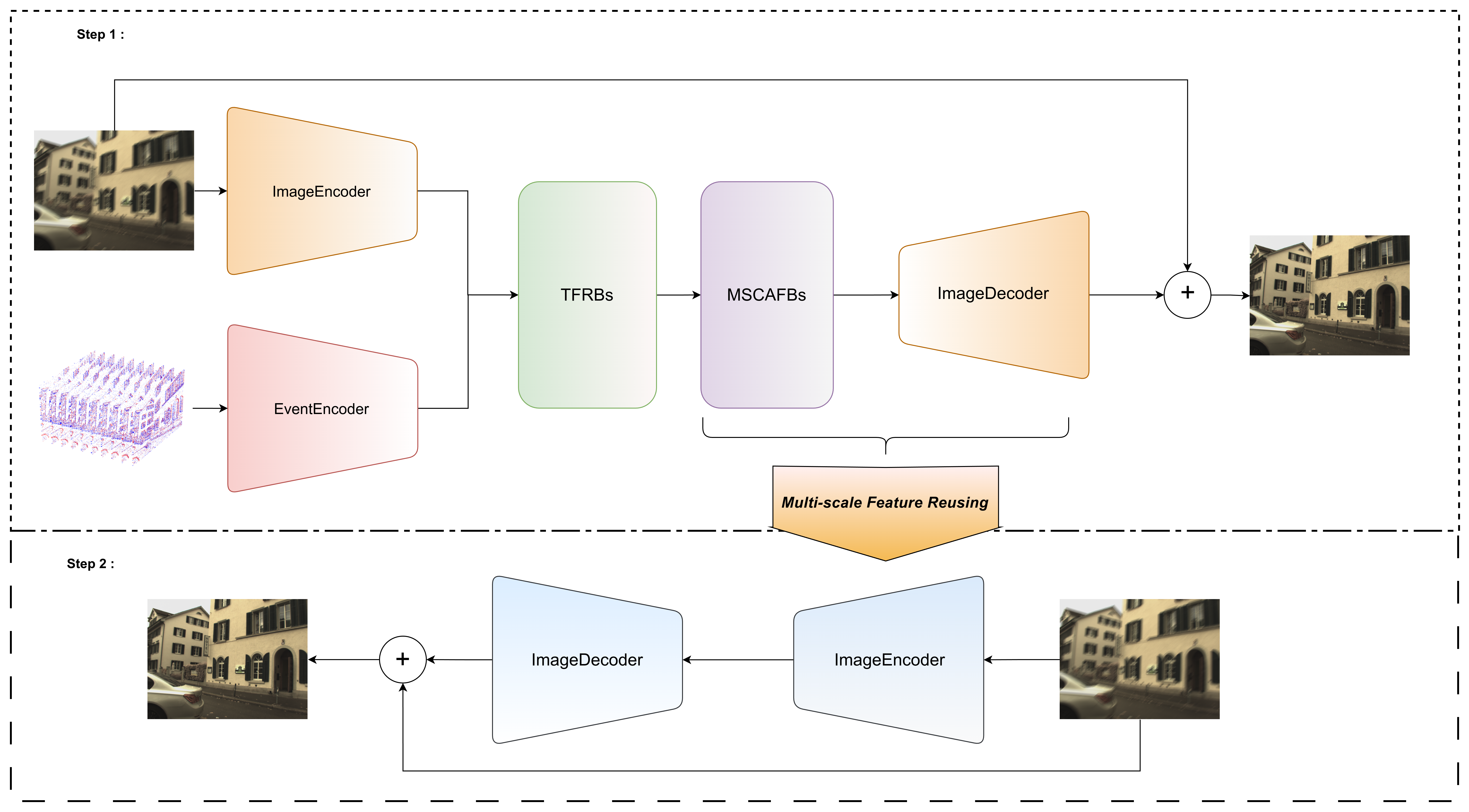} 
    \caption{The framework of DASTF-Net, proposed by Team MiVideoDeblur.}
    \label{fig:model}
\end{figure}

\textbf{Introduction.}
As illustrated in Fig.~\ref{fig:model}, their team proposed the Dual Attention Spatio-Temporal Fusion Network(DASTF-Net).
Motivated by EFNet~\cite{Event_Based_Fusion_for_Motion_Deblurring_with_Cross_modal_Attention}, their model employs a two-stage encoder-decoder architecture. Initially, two encoders separately extract multi-scale features from both the image and event data. Based on the EGACA module~\cite{event_based_frame_interpolation_with_ad_hoc_deblurring} and the FAF module~\cite{Motion_Aware_Event_Representation-driven_Image_Deblurring}, they have designed the Temporal Fusion Residual Block (TFRB) and Multi-Scale Cross-Attention Fusion Block (MSCAFB), which perform feature fusion in the temporal and spatial dimensions, respectively. By incorporating a dual-attention mechanism, these modules effectively enhance the model's performance. Following feature fusion, the fused features are fed into a Restormer~\cite{Restormer_Efficient_transformer_for_high-resolution_image_restoration}, which further leverages the feature information to improve the model's performance. 

\textbf{Training strategy.}
They employed a four-stage training strategy. In the first stage, the network was trained for 160k iterations using the PSNRLoss function. AdamW Optimizer was used, with an initial learning rate of 2e-4 and a cosine annealing learning rate schedule for updates. Subsequently, in the second stage, data augmentation techniques were introduced, which included adding random Gaussian noise and applying random scaling to the input data. Building upon the model from the first stage, the training continued for 80k iterations with an initial learning rate of 1e-4. For the third and fourth stages, the patch size was progressively increased from 256 to 320 and then to 480. The network was trained for 40k iterations in the third stage and 45k iterations in the fourth stage.
\subsection{404NotFound}

\begin{figure*}[ht]
    \centering
    \includegraphics[width=0.90\linewidth]{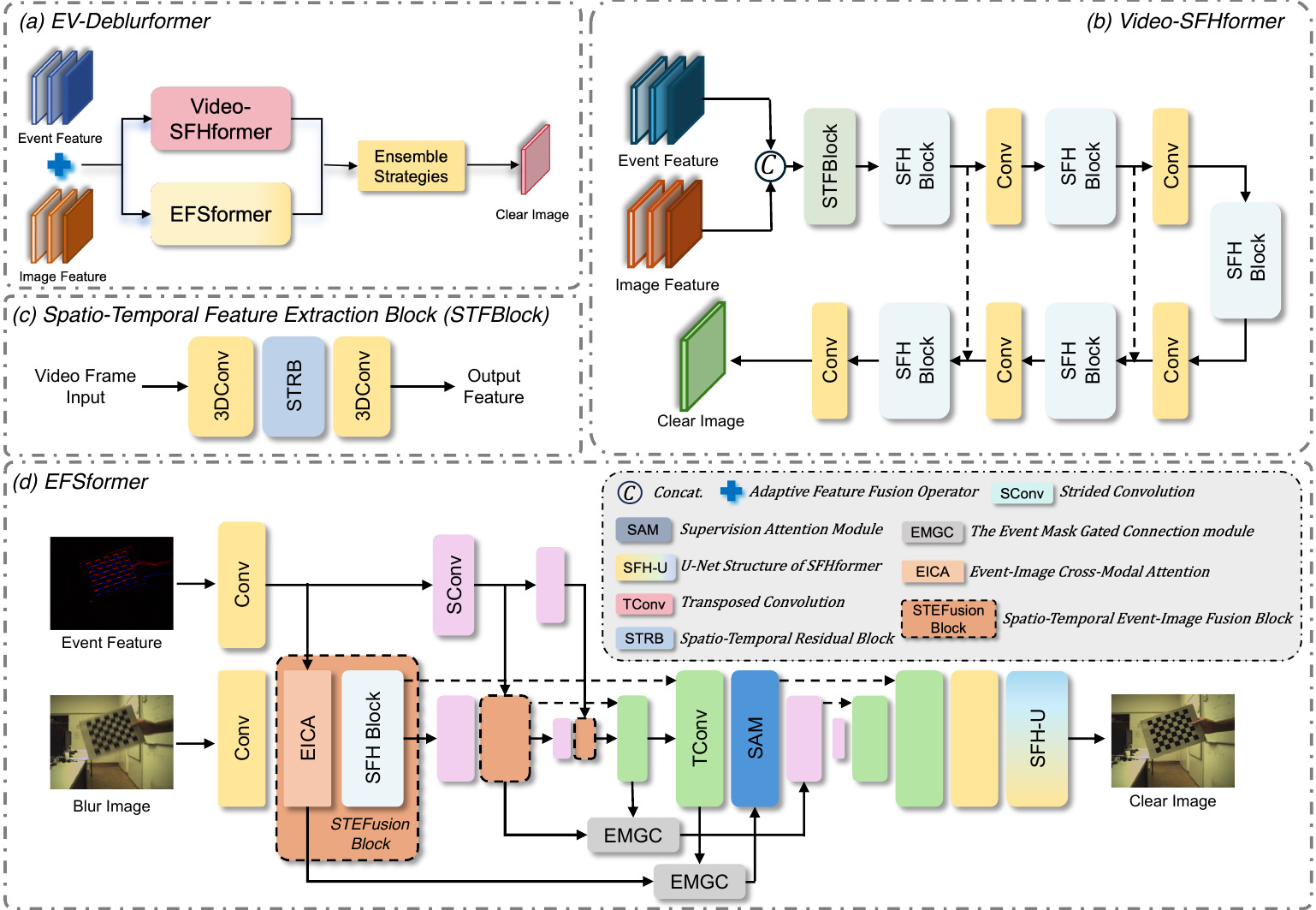}
    \caption{The architecture diagram of EV-Deblurformer, proposed by Team 404NotFound, is designed for event-guided motion deblurring.}
    \label{fig: Network Pipeline}
\end{figure*}

Their team proposes EV-Deblurformer\cite{16paper}, a framework consisting of two complementary models designed to fully leverage the temporal dynamics of video sequences and the rich texture details present in single images. The framework includes two distinct components: Video-SFHformer, developed for video-based deblurring, and EFSformer, tailored for single-image deblurring. In Video-SFHformer, they introduce STFBlock to enhance the model’s capacity for long-range temporal modeling. In EFSformer, they incorporate STEFusionBlock, which fuses event features from the frequency domain to improve spatial detail restoration. To achieve optimal performance, as shown in Section~\ref{sec: ensemble}, a sequence-level ensemble strategy is employed to merge the outputs of both models. A progressive training scheme is also adopted to enhance robustness and effectiveness.

\subsubsection{Overall Pipeline}

Figure~\ref{fig: Network Pipeline} illustrates the overall architecture of their proposed method, EV-Deblurformer. This approach, built upon the two models: Video SFHformer and EFSformer, fully exploits the rich temporal dynamics and sharp edge information provided by event data. For the video deblurring model, they propose the Video-SFHformer based on SFHformer. For the single-image motion deblurring model, they propose the EFSformer built on EFNet\cite{Event_Based_Fusion_for_Motion_Deblurring_with_Cross_modal_Attention}.

\subsubsection{Implementation Details}
They implement their proposed network via the PyTorch 2.1.2 platform. Adam optimizer with parameters $\beta_1=0.9$ and $\beta_2=0.999$ is adopted to optimize their network. Motivated by \cite{Restormer_Efficient_transformer_for_high-resolution_image_restoration} they introduce the progressive training strategy. The training phase of their network could be divided into two stages:

(1) \textbf{Initial training of EV-Deblurformer}.  They use a progressive training strategy at first. For the video-based motion deblurring model, they start training with patch size $152\times 152$ with batch size of 16 for 250K iterations. The patch size and batch size pairs are updated to $[(192^{2},12),(256^{2},8),(304^{2},8)]$ at iterations [250K, 200K, 150K]. The initial learning rate is $2\times 10^{-4}$ and remains unchanged when patch size is 192. Later, the learning rate is set to $1 \times 10^{-4}$ and $7 \times 10^{-5}$ for patch and batch size pairs of $(256^2,\ 8)$ and $(304^2,\ 8)$, respectively. They employ a cosine annealing learning rate decay strategy, gradually reducing the learning rate. For the single-image-based motion deblurring model,  They begin training with a patch size of $192 \times 192$ and a batch size of 12 for 250K iterations. During training, patch size and batch size pairs are progressively updated to $(256^2,\ 10)$, $(288^2,\ 8)$, and $(320^2,\ 8)$ at 36K, 24K, and 24K iterations, respectively. The initial learning rate is set to $5 \times 10^{-4}$, and later adjusted to $1 \times 10^{-4}$, $7 \times 10^{-5}$, and $5 \times 10^{-5}$ corresponding to the updated patch and batch size configurations. A cosine annealing schedule is employed to gradually decay the learning rate throughout the training process. The first stage is performed on the NVIDIA RTX 4090 GPU. They obtain the best model at this stage as the initialization of the second stage.

(2) \textbf{Fine-tuning EV-Deblurformer}. For the video-based motion deblurring model, they start training with a patch size of $320 \times 320$ and a batch size of 4 for 150K iterations. The initial learning rate is set to $1 \times 10^{-5}$ and is adjusted to $1 \times 10^{-7}$ using a cosine annealing schedule, over a total of 150K iterations. They use the entire training data from the challenge without applying any data augmentation techniques. The exponential moving average (EMA) is employed for the dynamic adjustment of the model parameters. For the single-image-based motion deblurring model, they adopt the same training strategy as used in the video-based motion deblurring model. The second training stage is conducted on an NVIDIA RTX 4090 GPU.

(3)\textbf{Evaluation Metrics}
They utilize two widely adopted reference-based evaluation metrics—Peak Signal-to-Noise Ratio (PSNR) and Structural Similarity Index Measure (SSIM)\cite{Image_qualityassessment:_From_errorvisibilitytostructural_similarity}—to evaluate the effectiveness of their method, following prior works\cite{chen2022simple, liang2021swinir, zamir2020learning, Restormer_Efficient_transformer_for_high-resolution_image_restoration}. Higher PSNR and SSIM values generally reflect better performance in image restoration tasks.

\subsubsection{Ensemble Strategies}
\label{sec: ensemble}
Ensemble learning has been proven to be an effective technique in image restoration. Its most basic application involves integrating the outputs of multiple models and applying a fusion strategy to achieve results with better generalization and greater stability in restoration quality.

The HighREV-test dataset consists of four sequences. Among them, one is an outdoor scene, which differs markedly from the other three in terms of object diversity, texture richness, and color composition. Based on this observation, they explore a sequence-level ensemble strategy that selectively exchanges outputs between Video-SFHformer and EFSformer.

Specifically, they start with the best-performing Video-SFHformer model and replace the output of the outdoor sequence in the HighREV-test set with the corresponding result generated by EFSformer. The results in Table~\ref{tab:rank} show that their approach yields the best performance, achieving the highest SSIM score and ranking third overall in the NTIRE Event-Based Image Deblurring Challenge.
\subsection{Give\_it\_a\_try}

\subsubsection{General method}

This submission is mainly based on the 
\href{https://github.com/eiffelcsy/efnet-new/}{public code of another team}. 
Models used in this submission are
EFNet att track fusion and EFNet att track fusion new,
which can be found at 
\href{https://github.com/eiffelcsy/efnet-new/tree/main/basicsr/models/archs}{archs}
or 
\href{https://github.com/eiffelcsy/efnet-new/tree/main/basicsr/models/archs/tested}{archs/tested}. They change
the training strategy, finetune the models and combine two
best models to push the limits of scoring.

\begin{itemize}
\item
How event modality is utilized in the deblurring process:
They used the given SCER format event voxels in training,
validating and training. The usage is as same as original
EFNet \cite{Event_Based_Fusion_for_Motion_Deblurring_with_Cross_modal_Attention} since new networks retain the encoder module
of the baseline.
\end{itemize}

\subsubsection{Implementation details}

\begin{itemize}
\item
Training:

In the first stage of training, all models are trained for
$2 \times 10^5$
iterations with a batch size of 16 by PSNR loss
function with AdamW optimizer. In each training batch,
each paired images and event voxel are randomly cropped to $256 \times 256$ and augmented by random flipping and rotation. The learning rate is initialized as $3 \times 10^{-4}$, and a cosine annealing scheduler is used to drop the final learning rate as $10^{-7}$. They finetuned the models obtained from the first stage with a patch size of $512 \times 512$. At this stage, all models are trained for another $2 \times 10^5$ iterations with a batch size of 4 and the learning rate drop from $2 \times 10^{-5}$ to $10^{-6}$. Models are validated for every $10^4$ iterations. Other settings remain unchanged.

\item 
Validating and Testing:

They chose the highest validated models for each network
during the fine-tuning stage and average two models’ output as final result to improve robustness.
\end{itemize}
\subsection{BUPTMM}

\begin{figure}[t]
    \centering
    \includegraphics[width=0.9\linewidth]{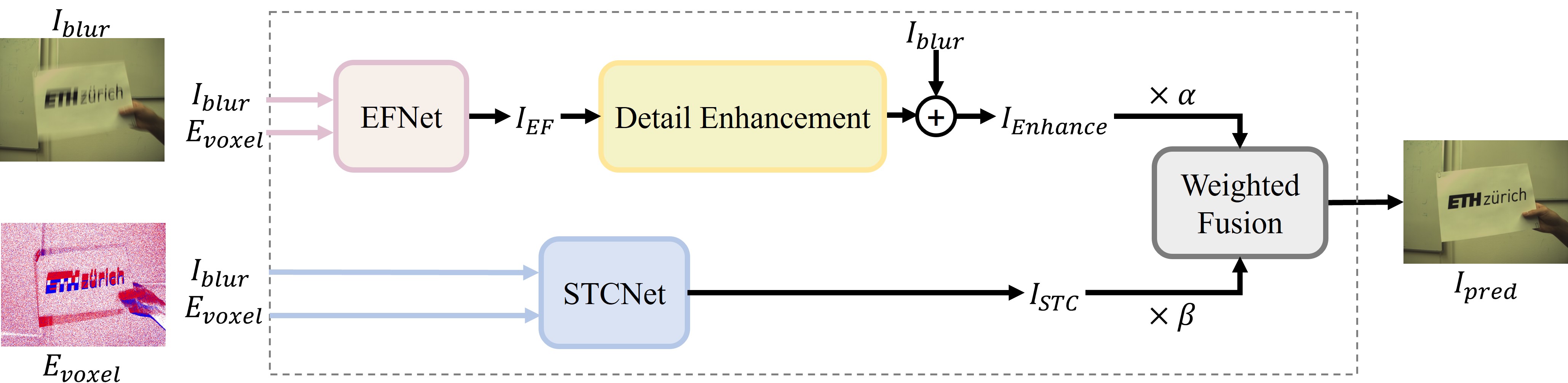}
    \caption{An overview of the method proposed by Team BUPTMM: They set the weights for the fusion, with $\alpha$ set to 0.6 and $\beta$ to 0.4.}
    \label{fig:buptmm_solution}
\end{figure}

\subsubsection{Architecture}
Our solution is built on EFNet\cite{Event_Based_Fusion_for_Motion_Deblurring_with_Cross_modal_Attention} and STCNet\cite{Motion_Deblurring_via_Spatial-Temporal_Collaboration_of_Frames_and_Events}. Inspired by \cite{Event-Based_Blurry_Frame_Interpolation_Under_Blind_Exposure}, they introduce a detail enhancement module that follows the EFNet prediction stage. The whole pipeline is illustrated in Fig.~\ref{fig:buptmm_solution}. The detail enhancement module adopts a simple U-Net structure.
\subsubsection{Implementation Details}
Both EFNet and STCNet are initialized with pre-trained GoPro checkpoints. They fine-tune them separately using the NTIRE official training dataset without additional data, aside from the pre-trained GoPro weights. The patch size is set to 1024×1024, and they employ the CosineAnnealingLR scheduler to adjust the learning rate.

The key differences in the training strategies for EFNet and STCNet are as follows:

For EFNet, they train EFNet for 100k iterations with a batch size of 4 using 4 NVIDIA H800 GPUs. The optimizer is AdamW with an initial learning rate of 2e-4. They generate the event voxel grid following the official script, setting the bin size to 24. Due to differences in the event encoder's channel size, they extended the pre-trained GoPro checkpoint weights from 6 to 24 bins. The loss function consists of the L1 loss, the Charbonnier loss, and the Sobel loss, with respective weights of 1.0, 0.5, and 0.5. Unlike the official EFNet implementation, they do not apply a mask between the two stages.

ForNet, they train STCNet for 1000 epochs with a batch size of 8 using 4 NVIDIA H800 GPUs. The optimizer is Adam with an initial learning rate of 2e-4. They use the official event voxel grid with a bin size of 6. The loss function is the Charbonnier loss.

\subsection{WEI}
\begin{figure*}[t]
    \centering 
    \includegraphics[width=0.95\textwidth]{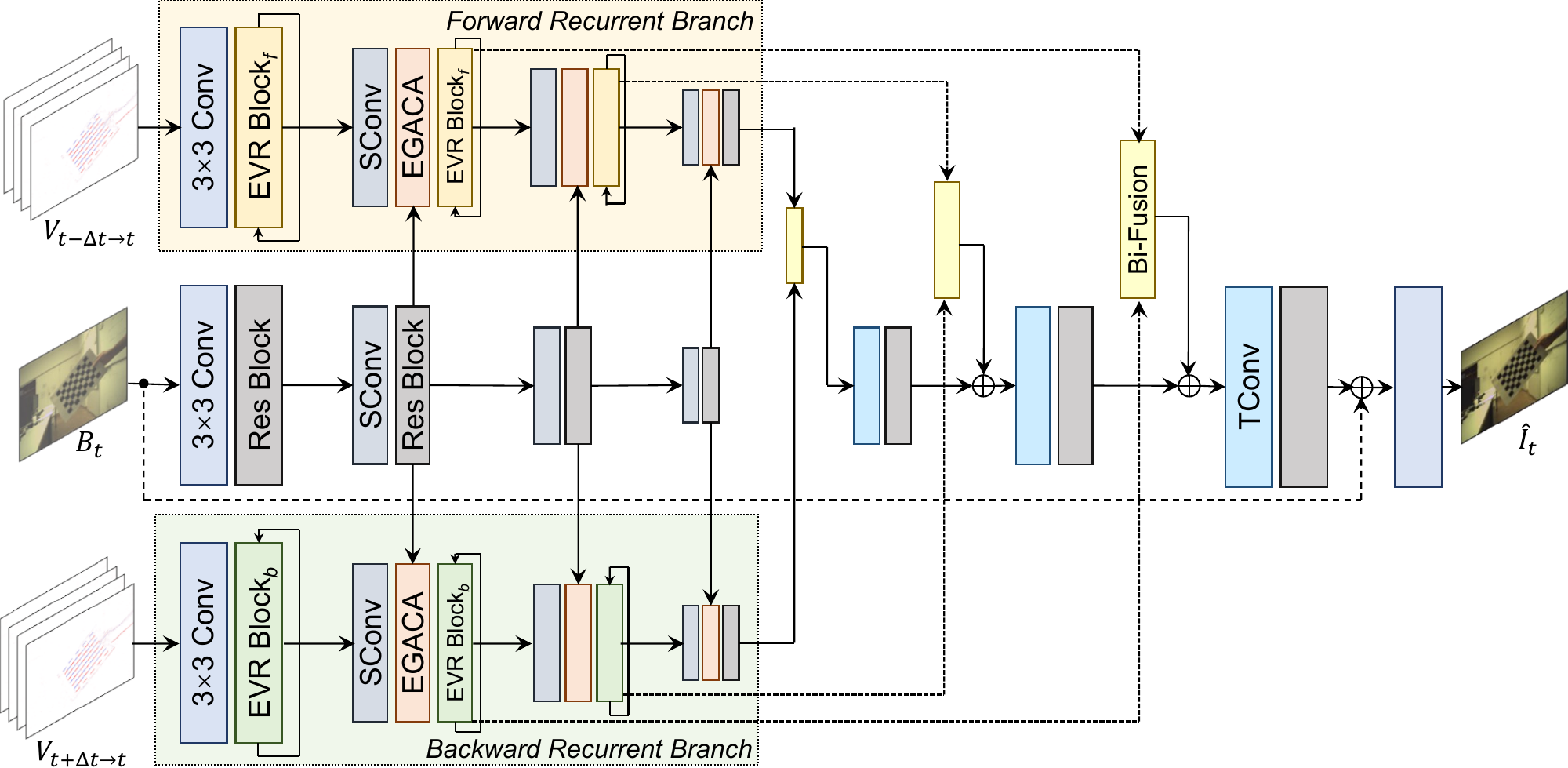}
    \caption{The architecture of the Bi-directional Gathered Recurrent Network (BGRN), proposed by Team Wei, is designed for event-based image deblurring and serves as an enhanced reconfiguration network for REFID.~\cite{event_based_frame_interpolation_with_ad_hoc_deblurring}. ``EVR Block'': event recurrent block~\cite{event_based_frame_interpolation_with_ad_hoc_deblurring}, ``EGACA'': event-guided adaptive channel attention~\cite{event_based_frame_interpolation_with_ad_hoc_deblurring}, ``SConv'': strided convolution, ``TConv'': transposed convolution, ``Bi-Fusion'': bidirectional fusion.}
    \label{fig:BGRN}
\end{figure*}

Since REFID~\cite{event_based_frame_interpolation_with_ad_hoc_deblurring} is an excellent method of event-based blurry video frame interpolation (VFI), considering the differences in modeling image deblurring and VFI problems, they adapt the REFID structure to fit the image deblurring challenge. As shown in Fig.~\ref{fig:BGRN}, they develop a Bi-directional Gathered Recurrent Network (BGRN) for event-based image deblurring.

\subsubsection{Network Architecture}
Following REFID~\cite{event_based_frame_interpolation_with_ad_hoc_deblurring}, the events within the exposure time ($t-\Delta t\rightarrow t+\Delta t$) are represented as a voxel grid $V_{t-\Delta t\rightarrow t+\Delta t}\in \mathbb{R}^{(M+1)\times H\times W}$, where $M$ is set to 9. Furthermore, they divide the voxel $V_{t-\Delta t\rightarrow t+\Delta t}$ into two segments $V_{t-\Delta t\rightarrow t}$ and $V_{t+\Delta t\rightarrow t}$ to perform forward and backward iterations, respectively.

The BGRN consists of image and event branches. Only a blurry image $B_t$ is fed into the image branch, and the network output is the corresponding sharp image $\hat{I}_t$. Besides, they split the original event branch into a forward recurrent branch and a backward recurrent branch, which respectively and recurrently consumes sub-voxels of forward event voxel $V_{t-\Delta t\rightarrow t}$ and backward event voxel $V_{t+\Delta t\rightarrow t}$ in a gathered way. In each recurrent iteration, the sub-voxel $V_{sub} \in \mathbb{R}^{2\times H\times W}$ is fed to the event branch, which encodes the event information for the latent frame. To fuse the features obtained from forward and backward recurrent branching, the outputs of both directions are fed into a channel cascade and $1 \times 1$ convolution at each scale (``Bi-Fusion'' in Fig.~\ref{fig:BGRN}). Then, they are added element by element with the features of the corresponding scale of the decoder. In addition, to reduce redundancy, they removed the recurrent structure of the decoder section and replaced it with residual blocks. Finally, to make the network learn high-frequency information, the output of the last residual block and the initial features of the blurred image are added element by element, and then the sharp image $\hat{I}_t$ is obtained through a $3 \times 3$ convolution.

\subsubsection{Implementation details}
\noindent \textbf{Training strategy.} They train BGRN with the HighREV training dataset specified by the organizer with a batch size of 4 for 200k iterations on an NVIDIA GeForce RTX 3090 GPU. They crop the input images and event voxels to $256 \times  256$ for training and use horizontal and vertical flips for data enhancement. AdamW~\cite{Decoupled_Weight_Decay_Regularization} with an initial learning rate of $2 \times 10^{-4}$ and a cosine learning rate annealing strategy with $1 \times 10^{-7}$ as the minimum learning rate are adopted for optimization. They use a PSNR loss~\cite{Event_Based_Fusion_for_Motion_Deblurring_with_Cross_modal_Attention} as supervision.

\noindent \textbf{Ensemble strategy.} During testing, they found that images prefixed with ``zigzag'' showed a large difference in brightness compared to other normal images. To adapt to this sudden change in brightness, they select images with the prefix ``sternwatz\_window'' similar to this scene from the training set. Then, they double their brightness to fine-tune the pre-trained BGRN model for 5k iterations with an initial learning rate of $2 \times 10^{-5}$. Therefore, the ensemble strategy is applied when testing, \textit{i.e.}, the abnormally bright images (prefixed with ``zigzag'') are processed with the fine-tuned model, and the others are processed with the initial pre-trained model.
\subsection{DVS-WHU}



\subsubsection{Network Architecture}

\begin{figure}[t]
    \centering
    \includegraphics[width=1.\linewidth]{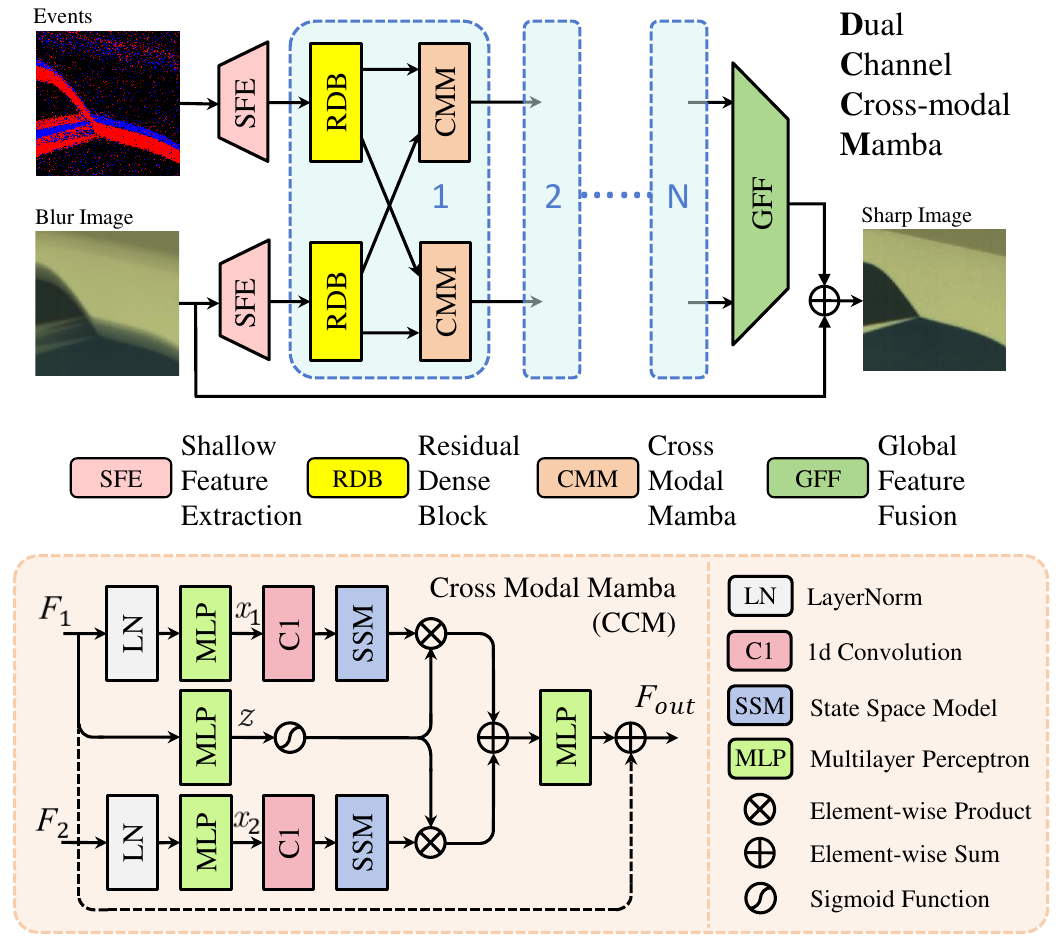}
    \caption{Architecture of DCCM, proposed by Team DVS-WHU.}
    \label{fig_dccm}
\end{figure}

 Positioned at \cref{fig_dccm}, the proposed Dual Channel Cross-modal Mamba (DCCM) architecture comprises three primary components: two Shallow Feature Extraction (SFE) modules, a series of \(N\) dual channel blocks (with \(N=20\) in their experimental configuration), each containing two Residual Dense Blocks (RDB) \cite{Residual_dense_network_for_image_super-resolution} and two Cross Modal Mamba (CMM) \cite{Pan-mamba_Effective_pan-sharpening_with_state_space_model} blocks, and a Global Feature Fusion (GFF) module. Initially, both blur image and events (represented in 24-bin voxel grids) are processed through the SFE module for preliminary feature extraction. Subsequently, the dual channel blocks facilitate in-depth feature extraction and cross-modal interaction. Finally, the GFF module synthesizes the ultimate latent sharp image.

The core concept of their network is to establish a mutual compensatory relationship between the features derived from event data and those from blurred images through a dual-channel framework. Specifically, while event data are often characterized by significant noise, images typically exhibit lower noise levels. The CMM block is employed to incorporate image features into the event data, thereby mitigating the noise present in the events. Conversely, event data are rich in sharp edge information, and the CMM block also facilitates the integration of event features into blurred images, ultimately contributing to the deblurred result. 

\subsubsection{Implementation Details}

The network is created with PyTorch and trained on two NVIDIA GeForce RTX 3090 GPUs for 150 epochs with ground-truth-guided L1 norm loss. The training process is composed of two phases. During the first phase, they follow the strategy of Cheng et al.\cite{Recovering_Continuous_Scene_Dynamics_from_A_Single_Blurry_Image_with_Events} and pretrain their DCCM on the mixed dataset including synthetic REDS dataset\cite{Ntire_2019_challenge_on_video_deblurring_and_super-resolution_Dataset_and_study} and semi-syhthetic HQF dataset\cite{Reducing_the_sim-to-real_gap_for_event_cameras} with a learning rate fixed at $1 \times 10^{-4}$ for 50 epochs. In the second phase, the network is fine-tuned on the HighREV dataset\cite{event_based_frame_interpolation_with_ad_hoc_deblurring} where the images are randomly cropped into $256 \times 256$ patches with horizontal flipping for data augmentation and the learning rate linearly decays to $1 \times 10^{-5}$ until the 150th epoch. 
\subsection{PixelRevive}

The model they used was the same as the EFNet\cite{Event_Based_Fusion_for_Motion_Deblurring_with_Cross_modal_Attention}. The key to the improved performance of their model lied in the utilization of additional datasets during training and the adoption of larger image sizes in the final fine-tuning phase. They employed a two-stage training strategy. First, they used an Events Simulator called V2E\cite{v2e_From_Video_Frames_to_Realistic_DVS_Events} to generate Events from REDS dataset. To generate the dataset, they used timestamp resolution as 0.001, dvs exposure duration as 0.001. The remaining parameters were configured identical to those specified in the V2E paper. They get over 20,000 pairs of events, blur images and sharp images. They trained the model on REDS for 250,000 iters, with gt\_size 256, patch size 8. When training on simulated datasets with the HighREV validation set, they observed a paradoxical divergence: while the training PSNR consistently improved, the validation PSNR exhibited a decline. This counterintuitive phenomenon may stem from distributional discrepancies between synthetic data and HighREV characteristics across multiple feature dimensions.

Then, they finetuned it on HighREV train dataset for 200,000 iters, with gt\_size 512, patch size 8. The TrueCosineAnnealingLR scheduler was employed in both training phases, configured with a period matching the total training iterations and a minimum learning rate value of 1e-7. After experiments, they found that larger gt\_size can improve the PSNR by about 0.5. Experiments showed performance decreases when gt\_size exceeds 512 (tested range: 256-608), making 512 the optimal size. Other strategy is same as the EFNet.
\subsection{CHD}
\begin{figure}[t]
    \centering
    \includegraphics[width=\linewidth]{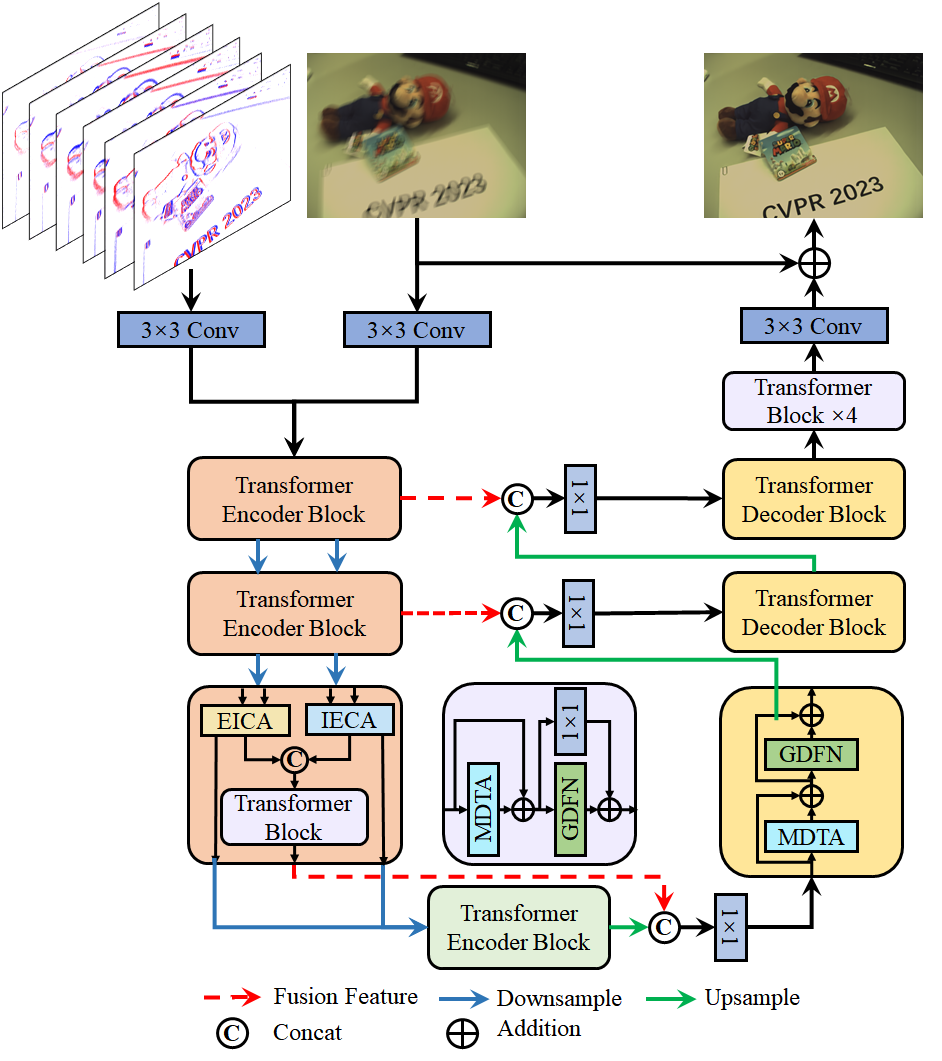}
    \caption{The framework of Event-Image Deblurformer Network (EIDFNet), proposed by Team CHD.}
    \label{fig:framework}
\end{figure}

As illustrated in Fig.~\ref{fig:framework}, team CHD develops an efficient Event-Image Deblurformer Network (EIDFNet) based on the Restormer architecture \cite{Restormer_Efficient_transformer_for_high-resolution_image_restoration}. To address the computational bottleneck encountered when restoring high-resolution blurry images using event data, they incorporate key design elements from EFNet \cite{Event_Based_Fusion_for_Motion_Deblurring_with_Cross_modal_Attention}.

\subsubsection{Network Architecture}
Considering the speed of model training, they still used the official 6-channel voxel grid event representation to achieve a balance between efficiency and precision. They input the blurred image and the event representation with consistent spatial resolution into the network and employ the modified Transformer Block to fuse the cross-modal feature. Firstly, they modify the transformer block in Restormer \cite{Restormer_Efficient_transformer_for_high-resolution_image_restoration} as a fusion module to achieve full interaction between different feature channels by setting the number of input and output dims in the GDFN and adding 1×1 convolution in the residual connections. Additionally, they build a mutually enhanced fusion encoder based on the Event-Image Cross-Modal Attention Fusion Module (EICA) proposed in EFNet \cite{Event_Based_Fusion_for_Motion_Deblurring_with_Cross_modal_Attention}. The enhanced image features are obtained using K and V derived from event embeddings, while Q is sourced from image embeddings. Conversely, the enhanced event features are generated with K and V originating from image embeddings, with Q being drawn from event embeddings. In order to achieve comprehensive integration of event and image features, the enhanced image features and enhanced event features are concatenated along the channel dimension. Subsequently, these concatenated features are fused using a Modified Transformer Block. Ultimately, each encoder produces enhanced image features, enhanced event features, and fused features. The enhanced event and image features undergo downsampling before being input into the subsequent encoder. The fusion feature is directly linked to the corresponding decoding feature through a skip connection.

\subsubsection{Training Strategy}
They perform progressive learning stategy fllow the settings in Restormer \cite{Restormer_Efficient_transformer_for_high-resolution_image_restoration} and trained the model on a A100 GPU with L1 loss. The network is trained on smaller image patches in the early epochs and on gradually larger patches in the later training epochs. During the training process, the batch sizes are [4,3,2,2,1,1], and the patch sizes are [128,160,192,256,320,384] with the iterations are [92000,64000,48000,36000,36000,24000]. They employ the AdamW optimizer with an initial learning rate 3e-4 that follows a CosineAnnealingRestartCyclicLR decay strategy.
\subsection{SMU}

\subsubsection{Motivation}
\label{sssec: motivation}

Inspired by recent successes in cross-knowledge sharing between events and RGB frames~\cite{Event_Based_Fusion_for_Motion_Deblurring_with_Cross_modal_Attention}, hierarchical temporal and frequency modelling~\cite{Frequency-aware_event-based_video_deblurring_for_real-world_motion_blur, event_based_frame_interpolation_with_ad_hoc_deblurring} and stage-wise fine-fusion~\cite{A_coarse-to-fine_fusion_network_for_event-based_image_deblurring} for the task of event-based RGB deblurring, they propose to modify the base EFNet model~\cite{Event_Based_Fusion_for_Motion_Deblurring_with_Cross_modal_Attention} such that the the modified model serves as a unified framework which (1) iteratively fine-tunes the coarser deblurred images through two stages of extensive fine-fusion to combat the insufficiencies of the existing decoding techniques while (2) can optionally be made to be specifically aware of propagated frequency information in latent representations to locally and globally filter the blur features in the RGB images through leveraging event features in the frequency domain. 

In addition, to the best knowledge, none of the existing methods for event-based RGB deblurring recognizes the importance of feature tracking in this task which can be beneficial especially in challenging conditions such as high contrast (i.e. very bright or dark surroundings) and fast motion (i.e., large pixel displacements within an accumulated event volume) scenarios~\cite{Data-driven_feature_tracking_for_event_cameras} towards robust performance. To address this limitation, they explicitly employ a data-driven feature tracking module in the pipeline, an inline feature tracker block, such that event feature tracks corresponding to different points in the reference RGB frame are intuitively incorporated in the learning process specifically in the initial stages of the unified framework. 

\begin{figure}[t]
    \centering
    \includegraphics[width=\linewidth]{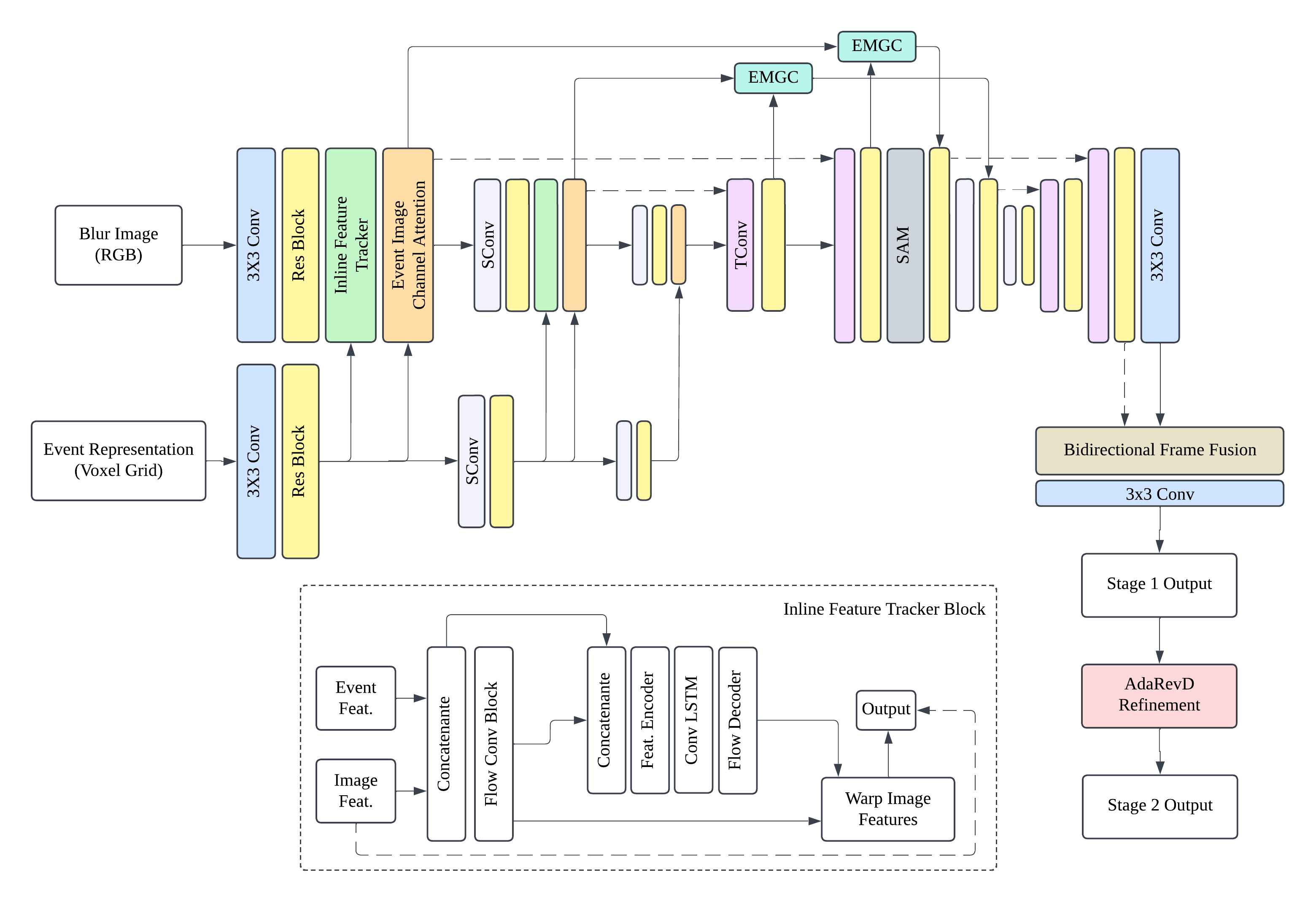}
    \caption{Overview of the proposed pipeline by Team SMU.}
    \label{fig:overview_smu}
\end{figure}

\subsubsection{Network Architecture}

As depicted in Fig.~\ref{fig:overview_smu}, they propose three main modifications: the inline feature tracker module, bidirectional frame fusion and AdaRevD refinement, to the original EFNet, backed by the motivation as described in section~\ref{sssec: motivation} and validated through the experiments. To this end, they design the inline feature tracker such that the latent RGB and event features are merged and learned through a flow autoencoder block in combination with a Conv-LSTM block to retrieve the temporal alignment of features. Furthermore, it is to be noted that they place the tracker at an initial stage of the pipeline to ensure that the tracker has the access to the high-level features of each modality, rather than the deeper low-level features, since high-level features, which are close to the input data, are more promising to contain information on temporal propagation, which is critical for co-aligned feature tracking. 

Inspired by~\cite{A_coarse-to-fine_fusion_network_for_event-based_image_deblurring}, they design the first stage of refinement using a bidirectional frame fusion block, specifically targeting the spatiotemporal information flow between adjacent coarse frames while in the second stage of refinement, they further refine the output from the first refinement stage with an objective to identify the still remaining degradation patterns in the RGB space and tackle them using an adaptive patch exiting reversible decoder module~\cite{Adarevd:_Adaptive_patch_exiting_reversible_decoder_pushes_the_limit_of_image_deblurring}. Optionally, to implement the frequency-based filtering of blur features, they follow the cross-modal frequency (CMF) module proposed by~\cite{Frequency-aware_event-based_video_deblurring_for_real-world_motion_blur} such that latent representations at each level of the first U-Net are passed through CMF modules, and concatenated in the decoder levels, in a hierarchical fashion to enhance the latent feature representations with frequency-aware characteristics.    

\subsubsection{Implementation Details}

They train the models using one NVIDIA 3090 GPU machine in two stages: (1) primary event-RGB fusion pipeline including the proposed frequency-aware module, explicit feature tracking and the first iteration of refinement based on the bidirectional frame fusion block and (2) second iteration of refinement based on AdaRevD framework~\cite{Adarevd:_Adaptive_patch_exiting_reversible_decoder_pushes_the_limit_of_image_deblurring}.

By following the baseline implementation~\cite{Event_Based_Fusion_for_Motion_Deblurring_with_Cross_modal_Attention}, they train the models on the HighREV dataset, in both stages, with an initial learning rate of $2\times10^{-4}$ for a total of $2 \times 10^{4}$ iterations. The utilized optimizer is AdamW~\cite{Decoupled_Weight_Decay_Regularization} and the learning objective is set to be PSNR loss~\cite{Event_Based_Fusion_for_Motion_Deblurring_with_Cross_modal_Attention}.  
\subsection{JNU620}

 \begin{figure*}[t]
    \centering
    \includegraphics[width=0.8\linewidth]{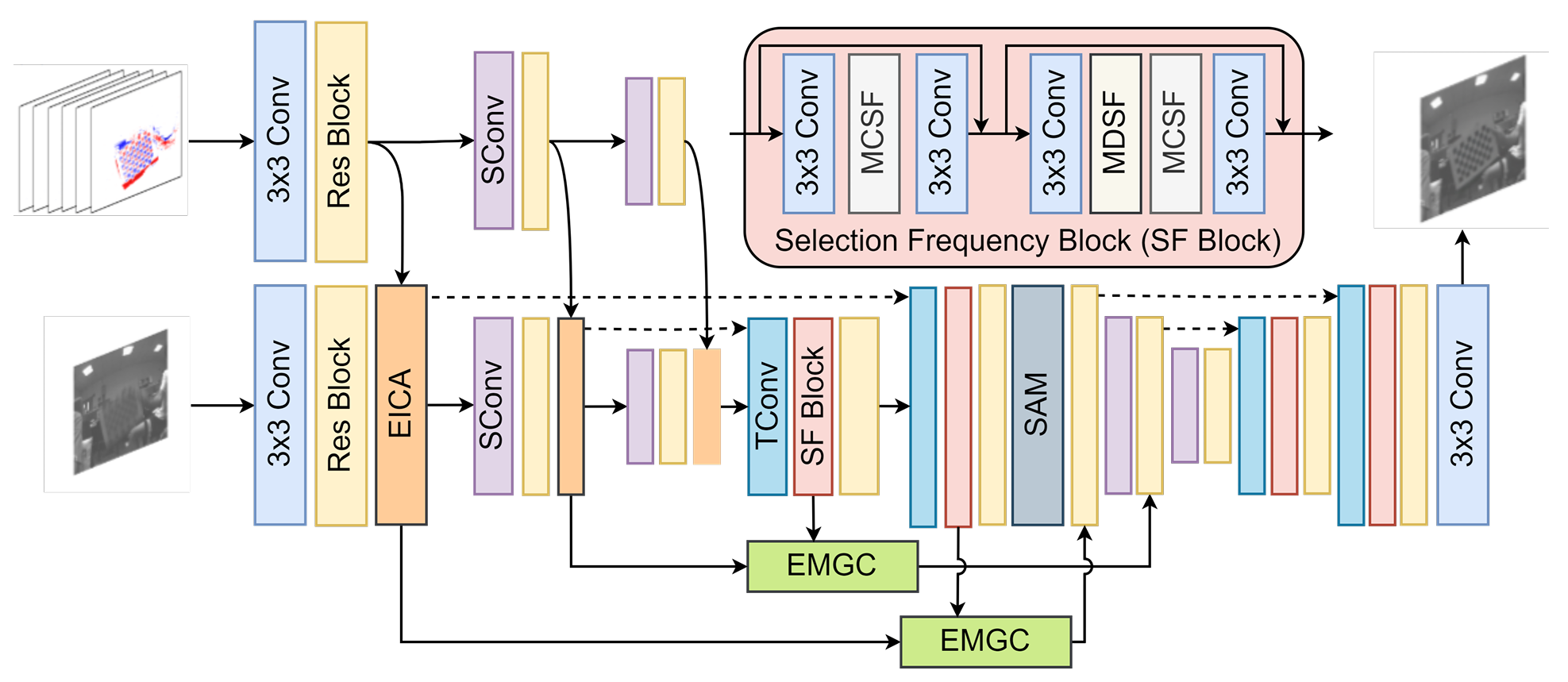}
    \caption{The model framework proposed by Team JNU620.}
    \label{JNU620}
\end{figure*}

 As shown in Fig. \ref{JNU620}, their framework adopts EFNet \cite{Event_Based_Fusion_for_Motion_Deblurring_with_Cross_modal_Attention} as the baseline architecture. To enchance frequency-aware feature
 processing, a selection frequency block (SF Block) \cite{cui2023selective} is
 integrated following each decoder. The architecture introduces two key components: 1) A multi-branch dynamic selection frequency (MDSF) module that adaptively decouples feature mappings into distinct frequency components
 through dynamic convolution operations; 2) A multi-branch compact selection frequency (MCSF) module specifically designed to expand the receptive field for processing degraded blurry images. Multiple data augmentation strategies were employed, including horizontal and vertical shiftings. For data preparation, they implemented multiple augmentation strategies including horizontal and vertical spatial shifts. The model was trained for 120,000 iterations on an NVIDIA GeForce RTX 3090 GPU with a batch size of 4. The models were optimized by the Adam method with $\beta_1$ = 0.9 and $\beta_2$ = 0.99 and the weight decay was set to $10^{-4}$. The initial learning rate was set to $2 \times 10^{-4}$, gradually decreased following a cosine annealing schedule. In inference phase, each test image undergoes augmentation through horizontal and vertival flips before input into the model. The final restored image is generated by averaging all augmented outputs.

\subsection{colab}
Our team proposes an improved method based on EFNet, named DEFNet (Dynamic Enhanced Fusion Network). This method incorporates three key enhancements. First, we introduce a multi-scale dynamic fusion module, which fuses event and image features at multiple spatial resolutions, significantly improving the restoration of fine details in blurred areas\cite{Event_based_video_deblurring_based_on_image_and_event_feature_fusion}. Second, we enhance the original EICA module by integrating a bidirectional attention mechanism, enabling more effective mutual guidance and interaction between image and event features. Third, for processing event data, we adopt a weighted interpolation strategy\cite{event_based_frame_interpolation_with_ad_hoc_deblurring} that models the dynamic weighting of event sequences more accurately, thereby enriching the temporal details provided to the image restoration process.
\subsubsection{Network}

\begin{figure}[t]
    \centering
    \includegraphics[width=0.7\linewidth]{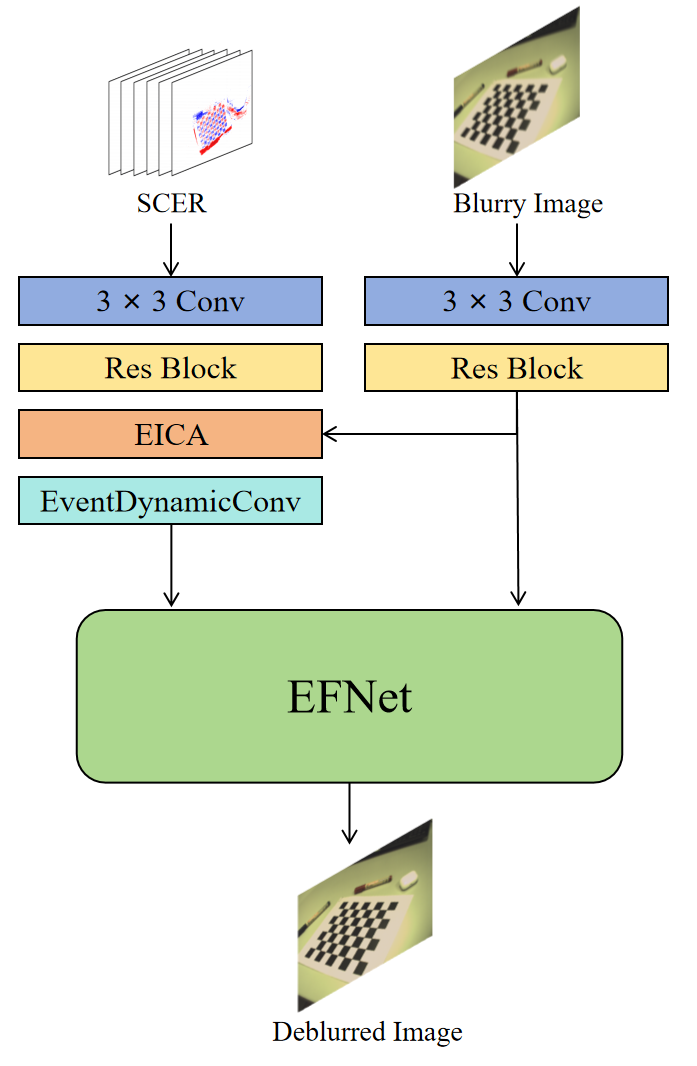}
    \caption{DEFNet architecture, proposed by Team colab.}
    \label{aaa}
\end{figure}

Fig. \ref{aaa} presents the architecture of DEFNet, which is built upon EFNet and incorporates the newly introduced modules: the multi-scale dynamic fusion module and the enhanced EICA module with a bidirectional attention mechanism. These components work collaboratively to optimize the motion deblurring process by improving feature representation and fusion between the image and event data.

During the deblurring process, event streams are used to provide fine-grained temporal variation information that guides the restoration of motion blur in image frames. Specifically, the Symmetric Cumulative Event Representation (SCER) encodes the temporal distribution of events while the enhanced Event-Image Cross-modal Attention Fusion (EICA) module leverages bidirectional attention to facilitate deeper interaction between modalities. Additionally, the integration of weighted interpolation improves the temporal alignment and accuracy of event feature extraction.Together, these components enable DEFNet to more effectively restore motion-blurred images by enhancing edge sharpness, preserving texture, and capturing motion dynamics with higher fidelity.

\subsubsection{Implementation Details}

We use the AdamW optimizer with an initial learning rate of 2e-4, weight decay of 1e-4, and betas set to [0.9, 0.99]. To dynamically adjust the learning rate, we used the TrueCosineAnnealingLR scheduler with a maximum iteration count of T\_max = 200000 and a minimum learning rate of 1e-7. During training, the batch size was set to 4, and 3 worker threads were used per GPU. The total number of training iterations was set to 40000.This method was trained and validated on the HighREV dataset. The model achieved significant improvements on both the training and validation sets, with PSNR and SSIM used as evaluation metrics during training. Validation was performed every 10,000 iterations, and the model was regularly saved.

\subsection{CMSL}

The Cascade Event Debluring Model With Event Edge Loss was build based on EFNet \cite{Event_Based_Fusion_for_Motion_Deblurring_with_Cross_modal_Attention}. An motion edge loss and a cascade framework were introduced to enhance the performance of EFNet.

\begin{figure}
    \centering
    \includegraphics[width=1.0\linewidth]{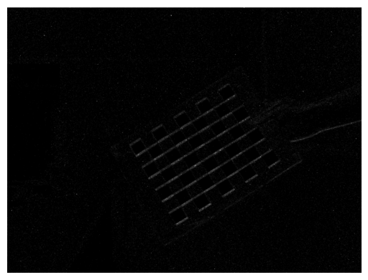}
    \caption{The visualization of the motion edges.}
    \label{motion edge}
\end{figure}

\begin{figure}
    \centering
    \includegraphics[width=1.0\linewidth]{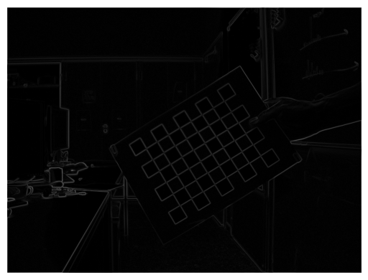}
    \caption{The edges in the ground truth frame}
    \label{true edge}
\end{figure}

The EFNet backbone was adopted and two improvements were proposed. Firstly, the event data were organized and represented as voxel \cite{Event_Based_Fusion_for_Motion_Deblurring_with_Cross_modal_Attention}. Then, two frame of the event voxels that were most close to the center of the exposure time were multiplied to produce a motion edge frame. The motion edge frame contains the edge of the moving objects in the current frame as shown in fig.~\ref{motion edge}, fig.~\ref{true edge} is the corresponding edge of the groud truth image (sharp image). As shown in fig.~\ref{motion edge} and fig.~\ref{true edge}, the motion edge contains clear lines that were consistent with the true edges and could served as a guiding information for image debluring. The edge of the deblured image output by the module should be similar to the motion edge. Therefore, a motion edge loss were proposed as follow:

\[{\ell _{edge}} = {\rm{mse}}(edge(\widehat x) \cdot m,e)\]
\[{m_{i,j}} = 1\quad{\rm{ if }}\quad {e_{i,j}}{\rm{  > }}\tau, \quad{\rm{ else\quad0}}\]
where mse(A,B) is the mean squared error between each element in matrix A and B, $\widehat x$ is the output deblured image, e is the motion edge frame, m is the motion edge mask, $\tau$ is the threthold parameter.

Secondly, a cascade frame work were proposed that two EFNet was connected in cascade to further enhance the image deblurring ability. The first EFNet took four frames of the event voxels that were relatively remote to the center of the exposure time while the second EFNet took two frames of the event voxels that were relatively close to the center of the exposure time. The two EFNet form a coarse-fine paradigm that gradually remove the motion delur.
\subsection{KUnet}


\subsubsection{Architecture}
Their solution is built upon a custom KUnet backbone tailored for event-based image deblurring. The model employs a dual-encoder strategy that separately processes RGB images and voxelized event data, each through a dedicated encoder branch. At the bottleneck, the features are fused via channel-wise concatenation and passed through a transformer module.

A key novelty in the design is the use of \textbf{KANLinear} layers within the transformer block. These layers, based on spline-interpolated kernels, improve attention expressiveness without adding significant computational overhead. This fusion architecture leverages the temporal sharpness of events with the spatial-semantic richness of RGB images to produce high-fidelity deblurred outputs.

\subsubsection{Implementation Details}
They train the model from scratch on the official NTIRE 2025 HighREV dataset without any external data or pretrained weights. The voxelized events are represented using 6 temporal bins, generating a 6-channel input tensor for the event encoder.

Training was conducted using 2 NVIDIA A100 GPUs with a batch size of 8 and a patch size of 256\(\times\)256. They trained the network for 150k iterations using the AdamW optimizer (\(\beta_1=0.9\), \(\beta_2=0.99\), weight decay = 1e-4) and a CosineAnnealingLR scheduler . Data augmentations included random horizontal flips and rotations.

The loss function includes a PSNR loss weighted at 0.5. Their final checkpoint achieved a peak PSNR of \textbf{29.42} on the NTIRE 2025 validation phase.

Inference was performed using a sliding window approach with a max minibatch size of 8. They observed an inference time of \(\sim0.15\) seconds per frame on an A100 GPU, and a memory footprint of approximately 16 GB during training.

\begin{figure}[t]
    \centering
    \includegraphics[width=0.45\linewidth]{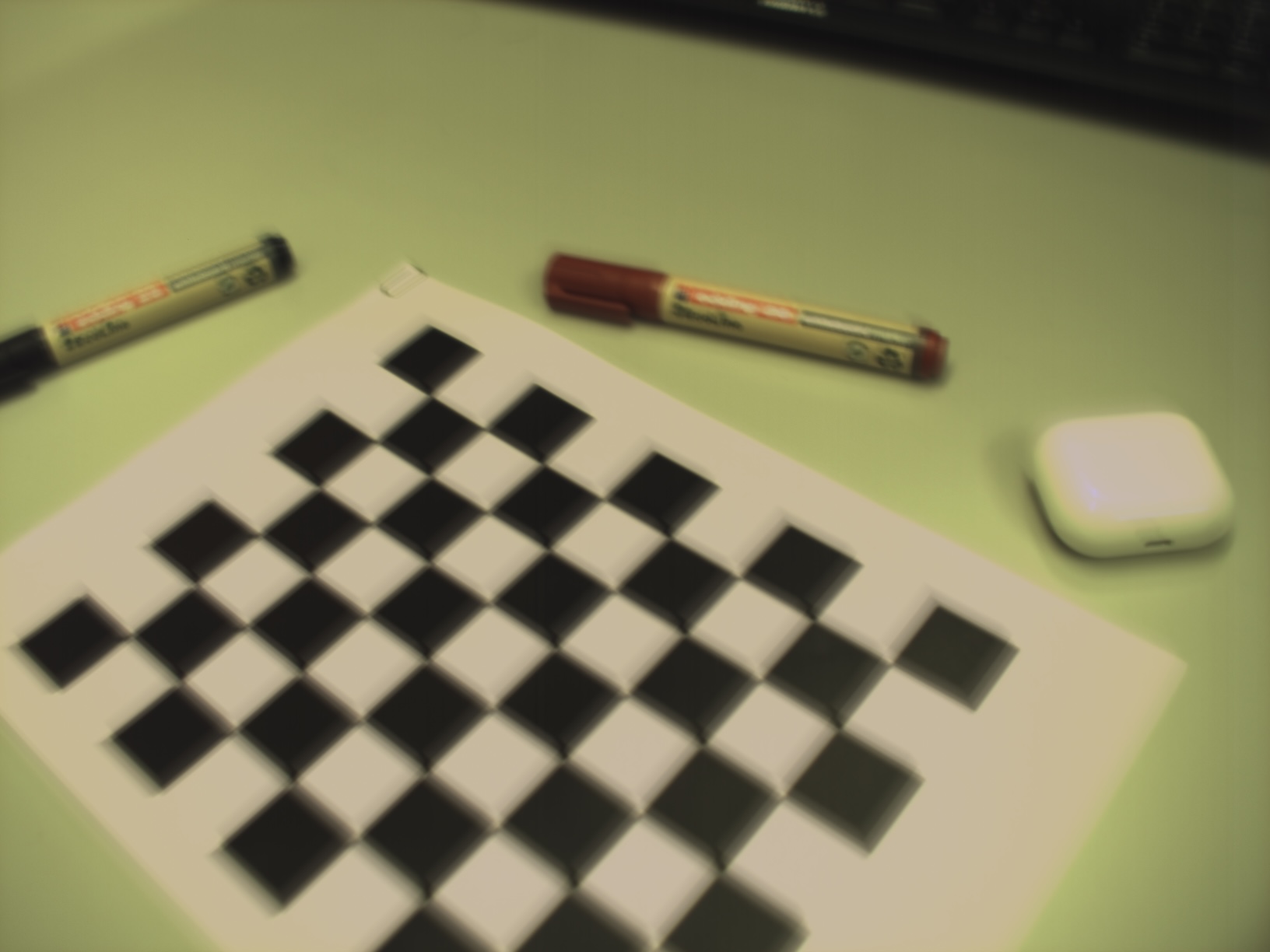}
    \includegraphics[width=0.45\linewidth]{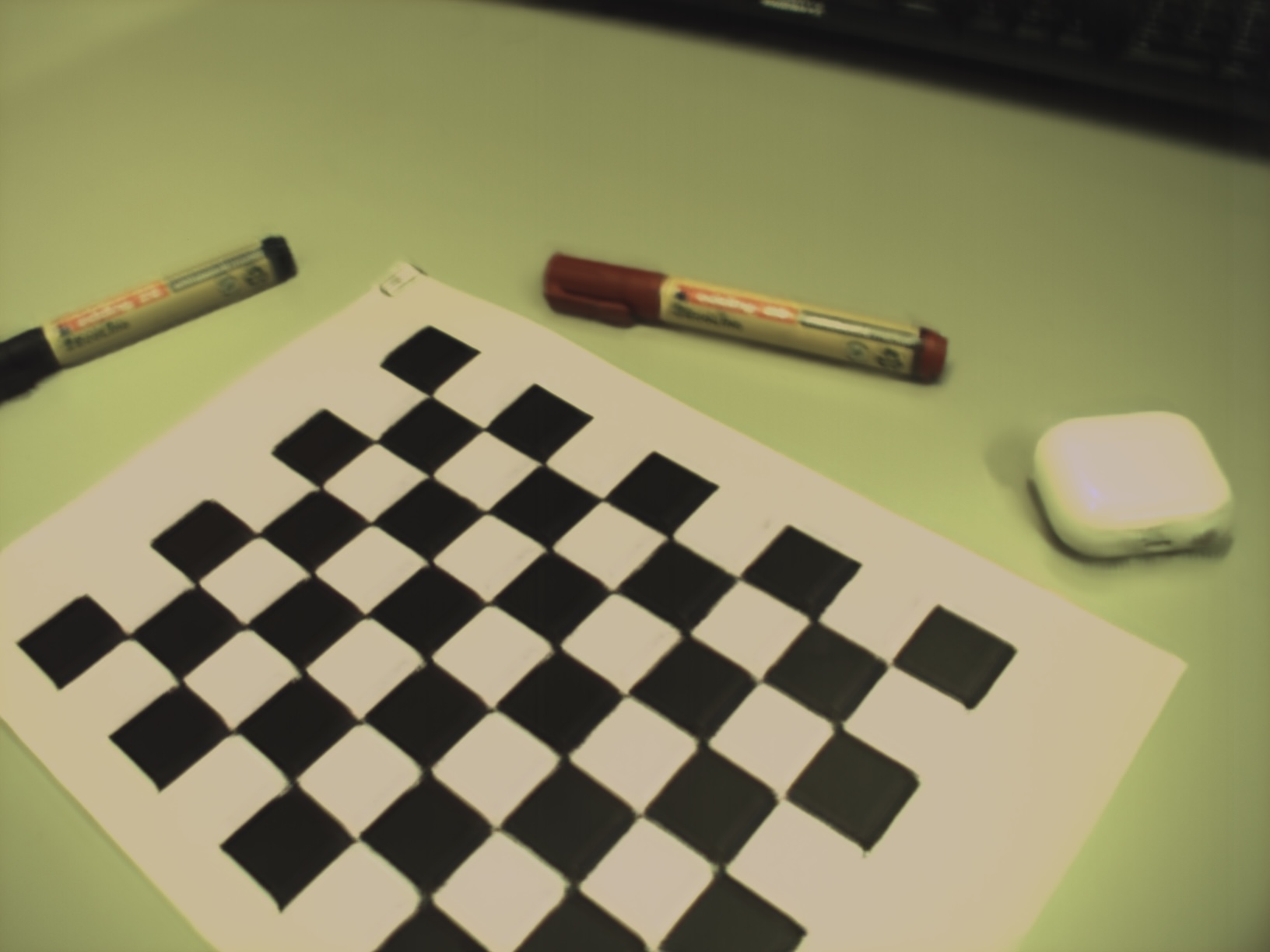}
    \caption{Left: Input blurry frame. Right: output of KUnet, with detailed texture.}
    \label{fig:kunet_result}
\end{figure}

\textbf{Model Complexity:}
\begin{itemize}
    \item Parameters: ~11M
    \item FLOPs: Not computed
    \item GPU Memory Usage: ~16 GB (training)
    \item Inference Time: ~0.15s/frame
\end{itemize}

\textbf{Code and Resources:}
\begin{itemize}
    \item GitHub: \url{https://github.com/Splendor73/NTIRE2025_EventDeblur_challenge_asu}
    \item Pretrained: \url{https://www.dropbox.com/scl/fi/l9td2xtbzxed2bg8tc9w0/17_KUnet.zip}
    \item Results: \url{https://www.dropbox.com/scl/fi/yrky29x2mdwt3k8e40yol/Results.zip}
\end{itemize}

\subsection{Group10}

The solution is built upon a custom adaptation of the EFNet deblurring framework\cite{Event_Based_Fusion_for_Motion_Deblurring_with_Cross_modal_Attention}. The method strategically harnesses both conventional image data and event-based information to mitigate motion blur effectively. Key components of the approach include:

     \noindent \textbf{Dual-Stream Network Architecture:} The model consists of parallel convolutional streams. One stream processes the blurry input image, while the other processes event data, which is converted into a voxel grid representation. A cross-modal attention module subsequently fuses the features extracted from both modalities, enhancing the network's ability to recover fine details in dynamic scenes.
    
    \noindent \textbf{Event Data Representation:} The raw event data -- comprising spatial coordinates, timestamps, and polarity -- is transformed into a voxel grid. This process involves temporal normalization and spatial mapping, enabling the network to capture the dynamic nature of motion events with high precision.
    
    \noindent \textbf{Training Strategy:} Utilizing mixed precision training to maximize GPU efficiency and accelerate the convergence process. Gradient accumulation is employed to effectively simulate a larger batch size, which is critical for stable training on high-resolution data. The training loss is computed using the Mean Squared Error (MSE) criterion, guiding the network to produce high-quality deblurred images.
    
    \noindent \textbf{Data Pipeline:} Custom PyTorch Dataset classes handle the loading and preprocessing of both image and event data. The pipeline includes resizing, normalization, and careful synchronization between blurry images and their corresponding event data, ensuring data consistency across modalities.
    
    \noindent \textbf{Performance Evaluation:} The evaluation strategy employs widely accepted metrics such as PSNR and SSIM to quantify restoration quality. Test outputs are resized to their original dimensions and saved as lossless PNG images to preserve the fidelity of the results.

Additional details include:

    \noindent \textbf{Parameter Count:} The EnhancedEFNet model consists of convolutional layers, CrossModalAttention blocks, and skip connections, leading to a parameter count in the range of millions.

    \noindent \textbf{CrossModalAttention layers:} These layers introduce additional tensor operations and memory usage. No external pre-trained models were directly used in training.The architecture was trained from scratch on the provided dataset.
    
    \noindent \textbf{GPU Memory Usage:} Memory usage is influenced by
    Batch Size, Default batch size of 4 per GPU, and
    Voxel Grid Representation, Uses 6 event bins, increasing input size.
    
    \noindent \textbf{CrossModalAttention:} Inspired by self-attention mechanisms in Transformer models. Hybrid Loss Function: Combines MSE and L1 loss for better generalization.CosineAnnealingLR Scheduler: Used to dynamically adjust learning rates during training.

    \noindent \textbf{Use of Additional Training Data: }Only NTIRE Dataset Used:The training was restricted to the HighREV dataset provided by NTIRE. No additional synthetic or external event-based datasets were incorporated. Potential Future Enhancements: Using real-world event datasets (e.g., DSEC, MVSEC) could improve generalization. Fine-tuning with pre-trained image restoration models (like DeblurGAN) could be explored.

     \noindent \textbf{Quantitative and Qualitative Improvements Quantitative Improvements (Metrics \& Performance): Peak Signal-to-Noise Ratio (PSNR):} Achieved PSNR: 25.93.Improved compared to baseline event fusion models. Structural Similarity Index (SSIM): Achieved SSIM: 0.82. Indicates better perceptual quality in restored images. Qualitative Improvements (Visual Results \& Generalization): Better Detail Recovery: The attention-based fusion of events and images leads to sharper edges and better contrast in reconstructed images. Works well in low-light or high-motion blur scenarios.

     \noindent \textbf{Comparison with Baseline Models:} Standard CNN-based deblurring struggles with fine-grained event details, but EnhancedEFNet effectively fuses event features to improve deblurring accuracy. CrossModalAttention aids in spatial alignment of events and images, reducing artifacts. Failure Cases \& Future Improvements: Highly blurred images with saturated event data can still cause artifacts. More robust fusion mechanisms (e.g., transformer-based approaches) could further enhance performance.

\section*{Acknowledgments}
This work was partially supported by the Humboldt Foundation, the Ministry of Education and Science of Bulgaria (support for INSAIT, part of the Bulgarian National Roadmap for Research Infrastructure). Shaolin Su was supported by the HORIZON MSCA Postdoctoral Fellowships funded by the European Union (project number 101152858). 
We thank the NTIRE 2025 sponsors: ByteDance, Meituan, Kuaishou, and University of Wurzburg (Computer Vision Lab).

\appendix

\section{Teams and affiliations}
\label{sec:teams}

\subsection*{NTIRE 2025 team}
\noindent\textit{\textbf{Title: }} NTIRE 2025 Event-Based Image Deblurring Challenge\\
\noindent\textit{\textbf{Members: }} \\
Lei Sun$^1$ (\href{mailto:leo\_sun@zju.edu.cn}{leo\_sun@zju.edu.cn}),\\
Andrea Alfarano$^1$ (\href{mailto:andrea.alfarano@insait.ai}{andrea.alfarano@insait.ai}),\\
Peiqi Duan$^2$ (\href{mailto:duanqi0001@pku.edu.cn}{duanqi0001@pku.edu.cn}),\\
Shaolin Su$^3$ (\href{mailto:shaolin@cvc.uab.cat}{shaolin@cvc.uab.cat}),\\
Kaiwei Wang$^4$ (\href{mailto:wangkaiwei@zju.edu.cn}{wangkaiwei@zju.edu.cn}),\\
Boxin Shi$^2$ (\href{mailto:shiboxin@pku.edu.cn}{shiboxin@pku.edu.cn}),\\
Radu Timofte$^5$ (\href{mailto:radu.timofte@uni-wuerzburg.de}{radu.timofte@uni-wuerzburg.de})\\
Danda Pani Paudel$^1$ (\href{mailto:danda.paudel@insait.ai}{danda.paudel@insait.ai}),\\
Luc Van Gool$^1$ (\href{mailto:vangool@vision.ee.ethz.ch}{vangool@vision.ee.ethz.ch}),\\

\noindent\textit{\textbf{Affiliations: }}\\
$^1$ INSAIT, Sofia University ``St. Kliment Ohridski'', Bulgaria\\
$^2$ Peking University, China\\
$^3$ Computer Vision Center, Spain\\
$^4$ Zhejiang University, China\\
$^5$ University of W\"urzburg, Germany\\

\subsection*{IVISLAB}
\noindent\textit{\textbf{Title: }} Triple Event-stream Image Deblurring Network\\
\noindent\textit{\textbf{Members: }} \\
Qinglin Liu$^1$ (\href{mailto:qlliu@hit.edu.cn}{qlliu@hit.edu.cn}),
Wei Yu$^2$, Xiaoqian Lv$^1$, Lu Yang$^3$, Shuigen Wang$^3$, Shengping Zhang$^1$, Xiangyang Ji$^2$ 

\noindent\textit{\textbf{Affiliations: }} \\ 
$^1$ Harbin Institute of Technology, Weihai \\
$^2$ Tsinghua University \\
$^3$ Raytron Technology Co., Ltd. \\
\subsection*{MiVideoDeblur}
\noindent\textit{\textbf{Title: }} Event-Based Image Deblurring from Team MiVideoDeblur\\
\noindent\textit{\textbf{Members: }} \\
Long Bao$^1$ (\href{mailto:baolong@xiaomi.com}{baolong@xiaomi.com}),
Yuqiang Yang$^1$, Jinao Song$^1$, Ziyi Wang$^1$, Shuang Wen$^1$, Heng Sun$^1$\\
\noindent\textit{\textbf{Affiliations: }} \\ 
$^1$ Xiaomi Inc., China \\
\subsection*{404NotFound}
\noindent\textit{\textbf{Title: }} Event-Conditioned Dual-Modal Fusion for Motion Deblurring\\ 
\noindent\textit{\textbf{Members: }} \\
Kean Liu$^1$ (\href{mailto:rickyliu@mail.ustc.edu.cn}{rickyliu@mail.ustc.edu.cn}),
Mingchen Zhong$^1$, Senyan Xu$^1$, Zhijing Sun$^1$, Jiaying Zhu$^1$, Chengjie Ge$^1$, Xingbo Wang$^1$, Yidi Liu$^1$, Xin Lu$^1$, Xueyang Fu$^1$, Zheng-Jun Zha$^1$\\
\noindent\textit{\textbf{Affiliations: }} \\ 
$^1$ University of Science and Technology of China \\
\subsection*{Give\_it\_a\_try}
\noindent\textit{\textbf{Title:}}  Event-Based Image Deblurring from Team Give\_it\_a\_try\\
\noindent\textit{\textbf{Members: }} \\
 Dawei Fan$^1$ (\href{mailto:dawei.fan@partner.samsung.com}{dawei.fan@partner.samsung.com}),
Dafeng Zhang$^1$, 
Yong Yang$^1$\\
\noindent\textit{\textbf{Affiliations: }} \\ 
$^1$ Samsung Research China- Beijing (SRC-B)\\

\subsection*{BUPTMM}
\noindent\textit{\textbf{Title: }} Weighted Fusion for Event-based Image Deblurring\\
\noindent\textit{\textbf{Members: }} \\
Siru Zhang$^1$ (\href{mailto:zhangsr@bupt.edu.cn}{zhangsr@bupt.edu.cn}),
Qinghua Yang$^1$, Hao Kang$^1$, Huiyuan Fu$^1$, Heng Zhang$^2$, Hongyuan Yu$^2$, Zhijuan Huang$^2$\\
\noindent\textit{\textbf{Affiliations: }} \\
$^1$ Beijing University of Posts and Telecommunications, Beijing, China.\\
$^2$ Xiaomi Inc., China. \\
\subsection*{WEI}
\noindent\textit{\textbf{Title: }} Bi-directional Gathered Recurrent Network for Event-based Image Deblurring\\
\noindent\textit{\textbf{Members: }} \\
Shuoyan Wei$^1$ (\href{shuoyan.wei@bjtu.edu.cn}{shuoyan.wei@bjtu.edu.cn}),\\
Feng Li$^2$,
Runmin Cong$^3$\\
\noindent\textit{\textbf{Affiliations: }} \\ 
$^1$ Institute of Information Science, Beijing Jiaotong University \\
$^2$ School of Computer Science and Engineering, Hefei University of Technology \\
$^3$ School of Control Science and Engineering, Shandong University \\
\subsection*{DVS-WHU}
\noindent\textit{\textbf{Title: }} Dual Channel Cross-modal Mamba for Event-based Motion Deblurring\\
\noindent\textit{\textbf{Members: }} \\
Weiqi Luo$^1$ (\href{2020302121081@whu.edu.cn}{wikyluo@whu.edu.cn}),\\
Mingyun Lin$^1$, 
Chenxu Jiang$^1$,
Hongyi Liu$^1$,
Lei Yu$^2$
\noindent\textit{\textbf{Affiliations: }} \\ 
$^1$ School of Electronic Information, Wuhan University \\
$^2$ School of Artificial Intelligence, Wuhan University \\
\subsection*{PixelRevive}
\noindent\textit{\textbf{Title: }} Event-Based Image Deblurring from Team PixelRevive\\
\noindent\textit{\textbf{Members: }} \\
Weilun Li$^1$ (\href{mailto:member1@member1.com}{xyj961011@163.com}),
Jiajun Zhai$^1$,
Tingting Lin$^1$\\
\noindent\textit{\textbf{Affiliations: }} \\ 
$^1$  College of Optical Science and Engineering, Zhejiang University\\

\subsection*{CHD}
\noindent\textit{\textbf{Title: }} Event-Image Deblurformer Network\\
\noindent\textit{\textbf{Members: }} \\
Shuang Ma$^1$ (\href{mailto:member1@member1.com}{3125508679@qq.com}),
Sai Zhou$^2$, Zhanwen Liu$^3$, Yang Wang$^4$\\
\noindent\textit{\textbf{Affiliations: }} \\ 
$^1$ Chang'an University, Xi'an, China \\
\subsection*{SMU}
\noindent\textit{\textbf{Title: }} Explicit Feature Tracking and Iterative Refinement for Enhancing Event-based Image Deblurring\\
\noindent\textit{\textbf{Members: }} \\
Eiffel Chong$^1$,
Nuwan Bandara$^1$, Thivya Kandappu$^1$ (\href{mailto:thivyak@smu.edu.sg}{thivyak@smu.edu.sg}), Archan Misra$^1$\\
\noindent\textit{\textbf{Affiliations: }} \\ 
$^1$ Singapore Management University \\
\subsection*{JNU620}
\noindent\textit{\textbf{Title: }} Event-Based Image Deblurring from Team JNU620\\
\noindent\textit{\textbf{Members: }} \\
Yihang Chen$^1$ (\href{mailto:Ehang@stu.jnu.edu.cn}{Ehang@stu.jnu.edu.cn}),\\
Zhan Li$^1$,
Weijun Yuan$^1$,
Wenzhuo Wang$^1$,
Boyang Yao$^1$,
Zhanglu Chen$^1$\\
\noindent\textit{\textbf{Affiliations: }} \\ 
$^1$ Department of Computer Science, Jinan University, Guangzhou, China\\

\subsection*{colab}
\noindent\textit{\textbf{Title: }} Dynamic Enhanced Fusion Network for Event-based Image Deblurring\\
\noindent\textit{\textbf{Members: }} \\
Yijing Sun$^1$ (\href{mailto:member1@member1.com}{syj3508852939@163.com}),
Tianjiao Wan$^1$, Zijian Gao$^1$, Qisheng Xu$^1$, Kele Xu$^1$\\
\noindent\textit{\textbf{Affiliations: }} \\ 
$^1$ National University of Defense Technology \\
\subsection*{CMSL}
\noindent\textit{\textbf{Title: }} Cascade Event Debluring Model With Event Edge Loss\\
\noindent\textit{\textbf{Members: }} \\
Yukun Zhang$^1$ (\href{mailto:member1@member1.com}{zhangyukun@cmhi.chinamobile.com}),
Yu He$^1$, Xiaoyan Xie$^1$, Tao Fu$^1$\\
\noindent\textit{\textbf{Affiliations: }} \\ 
$^1$ China Mobile (Hangzhou) Information Technology Co., Ltd, Hangzhou, China\\
\subsection*{KUnet}
\noindent\textit{\textbf{Title}} KUnet\\
\noindent\textit{\textbf{Members: }} \\
Yashu Gautamkumar Patel$^1$ (\href{mailto:ypatel37@asu.edu}{ypatel37@asu.edu}),
Vihar Ramesh Jain$^1$,
Divesh Basina$^1$ \\
\noindent\textit{\textbf{Affiliations: }} \\ 
$^1$ Arizona State University \\
\subsection*{Group10}
\noindent\textit{\textbf{Title: }} Event-Based Image Deblurring from Team Group10\\
\noindent\textit{\textbf{Members: }} \\
Rishik Ashili$^1$ (\href{mailto:rishik67_soe@jnu.ac.in}{rishik67\_soe@jnu.ac.in}), Manish Kumar Manjhi$^1$, Sourav Kumar$^1$, Prinon Benny$^1$, Himanshu Ghunawat$^1$, B Sri Sairam Gautam$^1$, Anett Varghese$^1$, Abhishek Yadav$^1$\\
\noindent\textit{\textbf{Affiliations: }} \\ 
$^1$ Jawaharlal Nehru University, New Delhi, India\\

{\small
\bibliographystyle{ieeenat_fullname}
\bibliography{main}
}

\end{document}